\documentclass[acmsmall]{acmart}

\usepackage{makecell}
\usepackage{multirow}
\usepackage{adjustbox}
\usepackage{enumitem}
\usepackage{tabularx}
\DeclareMathOperator*{\argmax}{arg\,max}

\usepackage{breakcites}
\usepackage{titlesec}

\titlespacing*{\section}
{0pt}{2mm}{2mm}
\titlespacing*{\subsection}
{0pt}{1.5mm}{1.5mm}
\titlespacing*{\subsubsection}
{0pt}{1mm}{1mm}

\AtBeginDocument{%
  \providecommand\BibTeX{{%
    \normalfont B\kern-0.5em{\scshape i\kern-0.25em b}\kern-0.8em\TeX}}}

\begin{document}

\title{Controllable Data Generation by Deep Learning: A Review}

\author{Shiyu Wang}
\authornote{Both authors contributed equally to this research.}
\email{shiyu.wang@emory.edu}
\affiliation{%
  \institution{Department of Biostatistics and Bioinformatics, Emory University}
  \country{USA}
}
\author{Yuanqi Du}
\authornotemark[1]
\email{ydu6@gmu.edu}
\affiliation{%
  \institution{Department of Computer Science, Cornell University}
  \country{USA}
}
\author{Xiaojie Guo}
\email{xguo7@gmu.edu}
\affiliation{%
  \institution{JD.COM Silicon Valley Research Center}
  \country{USA}
}
\author{Bo Pan}
\email{bo.pan@emory.edu}
\affiliation{%
  \institution{Department of Computer Science, Emory University}
  \country{USA}
}
% \author{Chen Ling}
% \email{chen.ling@emory.edu}
% \affiliation{%
%   \institution{Department of Computer Science, Emory University}
% \author{Zheng Zhang}
% \email{zheng.zhang@emory.edu}
% \affiliation{%
%   \institution{Department of Computer Science, Emory University}
\author{Zhaohui Qin}
\email{zhaohui.qin@emory.edu}
\affiliation{%
  \institution{Department of Biostatistics and Bioinformatics, Emory University}
  \country{USA}
}
\author{Liang Zhao}
\email{liang.zhao@emory.edu}
\affiliation{%
  \institution{Department of Computer Science, Emory University}
  \country{USA}
}

\renewcommand{\shortauthors}{Wang and Du, et al.}
 
\begin{abstract}
Designing and generating new data under targeted properties has been attracting various critical applications such as molecule design, image editing and speech synthesis. Traditional hand-crafted approaches heavily rely on expertise experience and intensive human efforts, yet still suffer from the insufficiency of scientific knowledge and low throughput to support effective and efficient data generation. Recently, the advancement of deep learning has created the opportunity for expressive methods to learn the underlying representation and properties of data. Such capability provides new ways of determining the mutual relationship between the structural patterns and functional properties of the data and leveraging such relationships to generate structural data, given the desired properties. This article is a systematic review that explains this promising research area, commonly known as controllable deep data generation. First, the article raises the potential challenges and provides preliminaries. Then the article formally defines controllable deep data generation, proposes a taxonomy on various techniques and summarizes the evaluation metrics in this specific domain. After that, the article introduces exciting applications of controllable deep data generation, experimentally analyzes and compares existing works. Finally, this article highlights the promising future directions of controllable deep data generation and identifies five potential challenges.
\end{abstract}




\maketitle

\section{Introduction}
\label{sec:intro}

Data generation is an important field that aims to capture the inherent distribution of data to generate similar yet new data. The field is a long-lasting, fast-growing, one with broad applications in critical fields such as molecule design~\citep{you2018graph, jin2018junction, de2018molgan}, image editing~\citep{gregor2015draw, mirza2014conditional, liu2019conditional}, text generation~\citep{ guo2018long, tambwekar2019controllable} and speech synthesis~\citep{schroder2001emotional, yang2018aag, habib2019semi}. Data generation requires exploring and manipulating the complex data structures which historically lead to high cost, intensive manpower, rich domain knowledge in large (and usually discrete) searching space. Partially because of this, traditional methods for data generation are customized to specific domains so that domain-specific heuristics and engineering can be more easily applied~\citep{lippow2007progress, vourkas2016emerging, yang1998models, fischer2001security}. For instance, the process of drug design, which is to generate new molecular structures, typically requires chemists to hand-craft candidate structures and then test for desired properties such as solubility and toxicity. Computational methods such as generic algorithms also may assist with combinatorial search for molecule structures by designing molecular mutation and crossover rules based on domain knowledge~\citep{jensen2019graph}. The molecule structure space is huge, however: for instance, the number of realistic drug-like molecules is estimated to be around $10^{33}$~\citep{polishchuk2013estimation}, and they pose considerable difficulties to search and identify the structure of interest. Moreover, in many domains such as neuroscience, circuit design, and protein structure, our domain knowledge is still quite limited and incomplete. The paucity of understanding on the data generative process limits our capability in reproducing and even creating new ones with desired properties. Another example is the logic circuit design which aims to output the desired schematics of the integrated circuit. The traditional circuit design is a rather complex process that relies on a large amount of mathematical modeling of the behavior of circuit elements based on characteristics of charge~\citep{uyemura1999cmos, vourkas2016emerging} and selecting appropriate materials for different circuit devices according to their properties~\citep{chen2016logic, vourkas2016emerging}. Noticeably, detailed reviews for traditional data generation techniques can be found individually in specific domains~\citep{huang2016coming, vourkas2016emerging, farahani2013review, fischer2001security}.

In the recent years, the advancement of deep learning provides new opportunities to tackle the aforementioned challenges in data generation. Deep learning techniques have exhibited significant success in learning the representation of various data types including images, texts, sequences and graphs~\citep{kan2022fbnetgen, ling2021deep, shrestha2019review, pouyanfar2018survey, pamina2019survey, wang2022multiobjective, zhang2021representation}. This success further empowers us to fit the mapping from data structures to their corresponding (latent) features, where the former can usually be discrete and unstructured while the latter are continuous vectors or matrices. Hence, instead of directly exploring the high-dimensional space of complex data structure with expensive combinatorial algorithms, the latent features of the data are explorable in the continuous vector space with efficient algorithms such as gradient-based ones. For instance, protein structure is formed by a sequence of amino acids and therefore the distribution of the sequential data can be captured and encoded by sequential deep learning models such as recurrent neural networks (RNNs) and transformer~\citep{muller2018recurrent}. Then the new sequence of amino acids can be autoregressively generated from the learned latent space of protein structure~\citep{ingraham2019generative}. Greater sequence diversity demonstrably has been achieved by those deep learning-based methods for protein design than with the traditional frameworks such as Rosetta~\citep{anand2022protein}. Since deep learning extracts the latent features in end-to-end fashion, moreover, the dependency on domain knowledge may be largely reduced. For example, in the field of image synthesis, deep learning-based techniques can learn the latent semantic representations of paintings of specific artists and readily fit the distribution in latent space and hence synthesizing new paintings of the same artists is simply a sampling$+$decoding process~\citep{hossain2019comprehensive}. Also because of this greater independence of domain knowledge, the data generation techniques based on deep learning could have better potential in being more easily generalized or cross-used in different data types or applications. 

Despite the promise of black-box deep learning techniques in tackling the traditional hurdles in data generation, how to fill the gaps between the learned latent features and the real-world properties of interest is critical to ensure the alignment between the generated data structures and the desired properties. Generating data with desired properties is de facto prerequisite in typical real-world applications, ranging from medicine design~\citep{xie2020mars, lambrinidis2021multi}, to circuit obfuscation~\citep{croft2022differentially}, art design~\citep{li2020abstract, mufti2020conditional}, and audio synthesis~\citep{tan2020generative, tits2019visualization}. For instance, chemists may not only generate novel quaternary ammonium compounds (QACs), but also hope that generated QACs have strong solubility in the water and minimum inhibitory concentration (MIC) smaller than 4mg$/$L to ensure antibiocity~\citep{buffet2011effect}. The community of image captioning may expect to generate more human-like texts from image with the length less than ten words in a humorous style~\citep{deng2020length, guo2019mscap}. As a result, to tackle this core crux in controlling the properties of the data generated by deep learning techniques, fast-growing demand, and research body in controllable deep data generation have been observed in recent years~\citep{guo2020property, you2018graph, chen2022fine, hu2017toward, liu2019conditional}. 

\subsection{Challenges in Controllable Deep Data Generation}
Despite its importance in making deep learning the de facto techniques for real-world, data-generation tasks, controllable deep data generation is technically nontrivial to achieve because of several major challenges:
\begin{itemize}[leftmargin=*]
\item\textbf{Large search space with property constraints}. The structure of the data to be generated, such as graph, image and text, can be highly complex, discrete and unstructured.Controllable deep data generation, therefore, needs to search in a large, discrete, and unstructured data space to identify the structure of data that satisfies desired property constraints. For instance, the number of realistic drug-like molecules is large and estimated to be around $10^{33}$~\citep{polishchuk2013estimation} and poses significant challenges on the efficiency of the generation model to search for the molecular structure that possesses desired properties. 

\item\textbf{Interference of multi-property control due to correlation among properties}. Correlation among properties of data is ubiquitous in many real-world settings. For instance, a scenery image of "nighttime" is more likely to have low "brightness". In addition, for molecules such as alkanes, increasing molecular size will likely result in a decrease in their water solubility~\citep{tolls2002aqueous}. Nevertheless, these correlated properties are difficult to control, since controlling one will constrain the others into some subspace such as a hyperplane or even a non-convex set, and that factor raises the need to design the corresponding optimization and generation strategies~\citep{hu2021causal, du2022interpretable}.

\item\textbf{Diversity in property types and heterogeneity in their control objectives}. The standard deep learning framework is typically designed for continuous data and trained via back-propagation. Nevertheless, discrete properties widely exist in real-world settings. For instance, linguists may want to generate a piece of audio speech with a specific emotion (e.g., anger, happiness, fear, sadness, neutral, etc.) while others also may want to synthesize a scenery image in the "morning". Difficulty to formularize and control different types of properties as multiple objectives, including continuous or discrete property values, property values within a range or maximizing (or minimizing) those properties, calls for various strategies and frameworks for controllable deep data generation.

\item\textbf{Efficiency issues in representation of complex data}. Deep data generation produces new data from the latent representation learned to cover the distribution of observed data. The structure of observed data can be rather complex, however. For example, the average length of proteins from eukaryote is 449, and each of these residues forms as any of 20 amino acids~\citep{zhang2000protein}. Complex molecules are typically made up of chains and rings that contain carbon, hydrogen, oxygen and nitrogen atoms and may have millions of those atoms linked together in specific arrays~\citep{201117}. As properties of data are usually highly sensitive to small changes of structure \citep{kirkpatrick2004chemical}, efficiently and expressively learning the representation of data is critical to generating new data with desired properties.

\item\textbf{Difficulty in labeling ground-truth data properties}. To train and evaluate the controllable deep data generation models, comparing the properties preserved by the generated data and the ground-truth properties is important. Although one can extract certain properties of generated molecules relying on well-developed domain knowledge and tools such as chemoinformatics toolkits~\citep{landrum2013rdkit}, this way is usually not the case for many domains where properties are difficult to be identified automatically for the data types such as text, audio, images, social networks, etc. For instance, the model will not know what style a piece of text preserves or what the emotion a specific audio expresses unless it is annotated either by human labor or by pre-trained models. 
\end{itemize}

\subsection{Our Contributions}
To date, a considerable amount of research has been devoted to controllable deep data generation to address the challenges. Understanding comprehensively the strengths and weaknesses of existing works is important for advancing the state-of-the-art techniques and foresee potential research opportunities. Also, broad interests exist in controlling data generation of various domains. Although most proposed methods have been designed targeting individual application domains, it is beneficial and possible to generalize their techniques to other application domains. Hence, cross-referencing these methods that serve different application domains is tough and needs to be addressed. The quality of results of controllable deep data generation. moreover, requires specifically designed evaluation strategies in various application domains. Consequently, systematic standardization and summarization of different evaluation strategies across various domains is necessary. In addition, given the growing demand from both Artificial Intelligence (AI) scientists who are looking for new available datasets to test their models for controllable deep data generation and domain-specific communities who are looking for more powerful controlling techniques to generate complex structured data with desired properties, the absence of a systematic survey of existing techniques for controllable deep data generation limits the advance of data generation for both parties. To fill the gap, this survey will provide a systematic review of controllable deep data generation techniques to help interdisciplinary researchers understand basic principles of controllable deep data generation, choose appropriate techniques to solve problems in their related domains and advance the research frontiers with standardized evaluation scenarios. The major contributions of this survey are as follows.

\begin{itemize}[leftmargin=*]
  \item \textbf{A systematic summarization, categorization, and comparison of existing techniques}. Existing techniques for controllable deep data generation are thoroughly categorized based on how the generation process is triggered to form a novel taxonomy of a generic framework. Technical details, advantages, and disadvantages among different subcategories of the taxonomy are discussed and compared. The taxonomy is proposed to enable researchers from various application domains to locate the technique that best fits their needs.
%   The techniques in the domain of controllable deep data generation are unified by proposing the taxonomy categorized by both methodologies and problem settings.
  \item \textbf{Standardized evaluation metrics and procedures}. Historically, data generation methods and their evaluations usually are customized to individual domains and are not well unified despite their shared abstract problems and goals. To tackle this challenge, the paper summarizes common evaluation metrics and procedures for controllable deep data generation and standardizes them from the perspective of both generated data quality and their property controllability. 
  \item \textbf{A comprehensive categorization and summarization of major applications}. Major applications including molecule synthesis and optimization, protein design, image editing and emotional speech generation are comprehensively introduced and summarized. A full comparison and discussion includes various techniques that have been applied to these application domains. The comprehensive categorization and summarization of these major applications will help the AI researchers explore extensive application domains and 
  +
  guide researchers in those domains to generate data with appropriate techniques.
  \item \textbf{Systematic review of existing benchmark datasets and empirical comparison of existing techniques on them}. Benchmark datasets borrowed in various application domains are systematically summarized according to various data modalities. In addition. experimental results are conducted by peer-reviewed articles and us to compare representative models for controllable deep data generation on those benchmark datasets. The systematic review of the existing benchmark datasets and the empirical comparison of representative techniques will enable model developers to involve additional datasets to evaluate their models and compare the performance of their proposed models with benchmark results. 
  
  \item \textbf{A helpful discussion of current research status and potential future directions}. Based on the comprehensive survey of various techniques for controllable deep data generation, standardized evaluation metrics, a broad spectrum of applications, systematic review of benchmark datasets and empirical comparisons of existing techniques, this paper concludes by offering profound insights into several open problems and outlining promising future directions in this domain.
\end{itemize}

\subsection{Related Works}
Previously published surveys have some relevancy to controllable deep data generation. These surveys can be classified into four topics: (1) deep learning, (2) deep representation learning, (3) deep data generation, (4) controllable data generation in specific domains. 

Deep learning is a general topic that covers a broad range of neural network-based techniques and includes two branches: (1) deep representation learning to embed complex data into expressive and informative low-dimensional representation and (2) deep data generation to recover the data from the low-dimensional representation. To date, the perspectives of deep learning have been thoroughly surveyed, including its history \citep{dong2021survey, pouyanfar2018survey, alom2019state, deng2014tutorial, hatcher2018survey}, objectives\citep{pouyanfar2018survey, liu2017survey}, models \citep{dong2021survey, pouyanfar2018survey, alom2019state, dargan2020survey, raghu2020survey, deng2014tutorial, schmidhuber2015deep, hatcher2018survey, liu2017survey} and so on~\citep{zhang2018survey, raghu2020survey}.

Deep representation learning has a huge impact on the downstream tasks, such as prediction \citep{butepage2017deep, lv2021anticancer}, classification \citep{tian2019multimodal, guo2017hybrid} and self-supervision~\citep{you2020does, ericsson2022self}. Deep representation learning has been surveyed from different perspectives, including history \citep{zhong2016overview}, motivation \citep{bengio2013representation}, learning strategies \citep{bengio2013representation, zhong2016overview}, models \citep{bengio2013representation, zhong2016overview}, training \citep{bengio2013representation, zhong2016overview}, challenges \citep{bengio2013representation} and future directions \citep{zhong2016overview}. Other surveys instead summarize representation learning techniques intended for various types of domains. For instance, Zhang et al.~\citep{zhang2018network} proposed new taxonomies to categorize and summarize the information network representation learning techniques, while Li and Pi~\cite{li2020network} reviewed a large number of network representation learning algorithms from two clear points of view of homogeneous network and heterogeneous network. Chen et al.~\cite{chen2020graph} reviewed a wide range of graph embedding techniques. Ridgeway~\cite{ridgeway2016survey} instead surveyed various constraints that encourage a learning algorithm to discover factorial representations that identify underlying independent causal factors of variation in data. Zhang et al.~\cite{zhang2021survey} continued to expand the topic by exploring the advancement of concept factorization-based representation learning ranging from shallow to deep or multilayer cases.

For deep data generation, Oussidi and Elhassouny~\cite{oussidi2018deep} gave an overview of the building blocks of deep generative models, such as variational autoencoders (VAEs) and generative adversarial networks (GANs), and pointed out issues that arise when trying to design and train deep generative architectures using shallow ones. Regenwetter et al.~\cite{regenwetter2022deep} further introduced key works that have introduced new techniques and methods, identified key challenges and limitations currently seen in deep generative models across design fields, such as design creativity and handling constraints and objectives. Furthermore, Eigenschink et al.~\cite{eigenschink2021deep} proposed a data-driven evaluation framework for generative models for synthetic data. Harshvardhan et al.~\cite{harshvardhan2020comprehensive} instead covered deep generation as part of generation in machine learning and proposed future directions. In addition to surveys on general deep data generation, other surveys may focus on the deep data generation in specific domains including graph generation \citep{guo2020systematic, zhu2022survey,faez2021deep}, image synthesis \citep{lu2017recent, luo2021survey}, text generation \citep{he2017deep,de2021survey} and audio generation \citep{cho2021survey, tan2021survey, mu2021review}. 

Surveys are available for controllable data generation techniques customized for specific domains. For instance, Ericsson et al.~\citep{huang2016coming} surveyed computational methodologies for \emph{de novo} protein design on the full sequence space guided by physical principles that underlie protein folding. Pederson~\citep{pederson1984historical} reviewed circuit analysis and its applications to computer-aided circuit design. Guihaire and Hao~\citep{guihaire2008transit} reviewed crucial strategic and tactical steps of transit network design. In addition to shallow models, deep models designed for controllable data generation were also reviewed in various domains. For instance, Toshevska and Gievska~\cite{toshevska2021review} claimed to have systemically reviewed text style transfer methodologies by deep learning, which covers controllable style generation on text data. Jin et al.~\cite{jin2022deep} is another survey that defined text style transfer aiming to control certain attributes in the generated text, such as politeness, emotion, humor, and many others. Another field related to controllable deep data generation is drug design, which usually controls properties of generated drug molecules~\cite{du2022molgensurvey}. Kim et al.~\cite{kim2021comprehensive} discussed the conditional control of generated compounds and the corresponding models for property control. Kell et al.~\cite{kell2020deep} instead only focused on the common-used model for drug design, VAEs, and provided examples of recent success of VAEs in terms of drug design. A recent effort by Gao et al.~\cite{gao2022sample} benchmarks molecule optimization task empowered by machine learning. In contrast, our survey reviews standardization, comparison, unification, categorization, and taxonomy of controllable deep data generation techniques spanning applications.

\subsection{Outline of the Survey}
We start the survey by first introducing the background of controllable deep data generation, challenges, our contributions, relationship between our survey and existing surveys in Section~\ref{sec:intro}. We next formally define the problem of controllable deep data generation (Section~\ref{sec:prob}) and propose the taxonomy according to various property control techniques (Section~\ref{sec:tax}) in Section~\ref{sec:ppt}, followed by the summarization of evaluation metrics in the same section. In Section~\ref{sec:tech}, we introduce techniques that have been developed for controllable deep data generation as well as detailed explanation of the concepts and representative works of controllable generation from scratch (Section~\ref{sec:gen}) and controllable transformation from source data (Section~\ref{sec:trans}) according to our taxonomy. In Section~\ref{sec:appl}, we showcase applications of models reviewed in Section~\ref{sec:gen} and Section~\ref{sec:trans} for various domain-specific tasks, followed by popular datasets employed in these domains. Later in Section~\ref{sec:chal}, we introduce the potential challenges, opportunities of the domain and limitations of the existing methods. We will conclude our survey in Section~\ref{sec:con}. In Appendix Section A, moreover, we introduce common frameworks for deep data generation as preliminaries. We also performed experimental comparison as well as analysis on popular molecular, image, text and audio datasets with common models for controllable deep data generation in Appendix Section B. 

\section{Problem Formulation, Taxonomy and Performance Evaluations}
\label{sec:ppt}
This section firstly formulates the problem of controllable deep data generation in Section~\ref{sec:prob}. Next we propose a taxonomy to summarize various techniques in this domain in Section~\ref{sec:tax}. We then unify standard metrics to evaluate the performance of these techniques for controllable deep data generation in Section~\ref{sec:eval}. 

\subsection{Problem Formulation}
\label{sec:prob}
We formally define the controllable deep data generation as follows. Controllable deep data generation aims to generate data with desired properties based on latent features extracted by deep learning techniques. Define $x\in\mathcal X$ as a data object such as an image, a graph, a text document, etc. Here $\mathcal X$ is the domain of the data. Also define $c$ as a set of $N$ properties of interest of $x$ such that $c=\{c_1, c_2,\cdots,c_N\}$, where each $c_i\ (i=1,\cdots,N)$ is the $i$-th property value of $x$. Moreover, we have $c^*=\{c^*_1, c^*_2,\cdots,c^*_N\}$ to hold the set of \textbf{desired} values (or ranges) of $N$ properties. In some situations, $c^*_i$ can be a binary value or real value. For example, when generating images of scenery, one may want to generate a nighttime (instead of daytime) image such that $c^*_i=\mbox{``nighttime''}$. In another example, one may want to generate a molecule with weight of 100 (amu) so that $c_i=$ 100. In some other situations, the desired property value $c^*_i$ can be a range or a set, for example, one may want to generate a molecule with ClogP\footnote{Here ClogP refers to the calculated value of partition coefficient (logP). logP of a material defines the ratio of its solubility in two immiscible solvents~\citep{box2008using}.} between 2 and 4. $c^*_i$ can have other forms depending on the applications. For example, when we do not have a specific desired value of the property but just want to maximize (or minimize) it, one could set it as (negative) infinite. Denote $z\in\mathbb{R}^{1\times K}$ as the latent representation of each data object $x$ learned by deep learning models. The distribution $p(z)$ characterizes the semantic distribution of data objects in latent feature space.

Controllable deep data generation aims to learn a data generator $g()$ that can generate data based on the latent semantic features $z\sim p(z)$ with its property values $c$ satisfying the user-desired $c^*$: $x\leftarrow g(z,x_0)\ \ \ \ s.t.,\ c\in c^*$, where $x_0$ is an optional term when any ``source data'' is available for new data object $x$ to derive and transfer from. For example, one can transfer a photo ($x_0$) into a painting of Van Gogh's style ($x$), or can edit a given molecule ($x_0$) toward a new one ($x$) with better properties. Such a scenario is called data transformation, which has wide, real-world applications and can be considered as an extension of data generation.

\subsection{Taxonomy}
\label{sec:tax}
\begin{figure}[h]
\begin{center}
\includegraphics[width=0.9\textwidth]{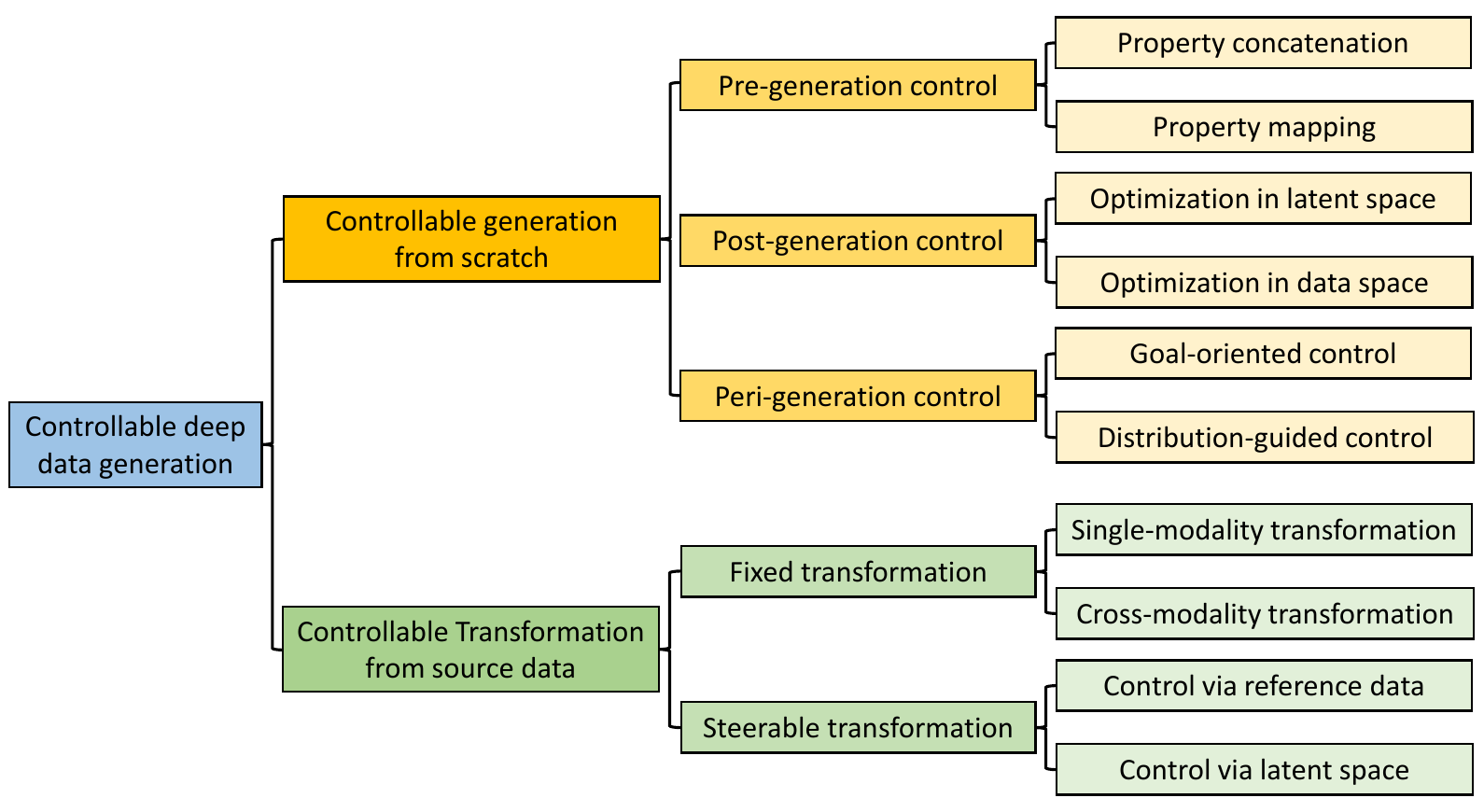}
% \vspace{-2mm}
\caption{The proposed taxonomy of controllable deep data generation.}
\label{fig:taxonomy}
% \vspace{-3mm}
\end{center}
\end{figure}

To have an overall view of this field and allow comparison of various techniques in a qualitative way, we propose taxonomy to categorize the existing works of controllable deep data generation techniques. Specifically, according to how the generation process is triggered, the proposed taxonomy has two main classes of works: 1) \textit{controllable generation from scratch} and 2) \textit{transformation from source data}. Specifically, controllable generation from scratch aims to generate data where specific attributes or features of the generated data can be controlled or guided by the user. Transformation from source data, by contrast, aims to transform from the data in the source domain to the data in the target domain while preserving target properties~\citep{guo2022graph}. Each class is further divided into subcategories based on how the property-control signal is introduced to the model. The detailed taxonomy including the subcategories of each aspect is presented in Fig. \ref{fig:taxonomy}. The comprehensive introduction of controllable generation from scratch and transformation from source data will be provided in Section~\ref{sec:gen} and Section~\ref{sec:trans}, respectively. In addition to the proposed taxonomy, we summarize the peer-reviewed publications following the aforementioned taxonomy in Table \ref{tab:data_summary}.

\begin{table}[h!]
    \centering
    \caption{Publications for each category of the taxonomy}
    \begin{adjustbox}{width=\textwidth}
    \begin{tabular}{|c|c|c|c|}
    \hline
    \multicolumn{3}{|l|}{Taxonomy} & Representative publications \\
    \hline\hline
    \multirow{6}{*}{\thead{Controllable generation\\ from scratch}} & \multirow{2}{*}{Pre-generation control} & Property concatenation & ~\cite{mirza2014conditional, kingma2014semi, cheung2014discovering,  odena2017conditional, hu2017toward, ficler2017controlling, esteban2017real, bodla2018semi,hoogeboom2022equivariant, nichol2021glide, henter2018deep, tevet2022human, rothchild2021c5t5,li2018learning,yang2021cpcgan,xu2019modeling, keskar2019ctrl, prabhumoye2020exploring,  tan2020generative, xu2020megatron,jin2020multi, mokhayeri2020cross, huang2022fastdiff} \\\cline{3-4}
                                &                                         & Property mapping &  ~\cite{jin2018junction,sohn2015learning,liu2019conditional,shao2021controllable,guo2020property,vasquez2019melnet,zang2020moflow,moon2020conditional,chen2020music, shoshan2021gan, lu2020structured, ardizzone2019guided,gomez2018automatic, choi2021ilvr}\\\cline{2-4}
                                & \multirow{2}{*}{Post-generation control} & Latent space optimization & ~\cite{ kang2018efficient, liu2021graphebm,deng2020disentangled, park2018data,mathieu2016disentangling, johnson2016composing, upchurch2017deep, kingma2018glow, camino2018generating, wang2020learning, ma2018constrained, goetschalckx2019ganalyze, mollenhoff2019flat, madhawa2019graphnvp, shen2021closed, du2022chemspace, guo2021deep, du2022disentangled, guo2021generating, pati2021disentanglement, li2021editvae, radford2015unsupervised,  du2020interpretable, zhou2019talking, guo2020interpretable, du2021deep, du2022small,  plumerault2020controlling, locatello2019disentangling, jahanian2019steerability, harkonen2020ganspace,  yang2020dsm, xu2020variational, tan2020music, parthasarathy2020controlled, wang2022deep}\\\cline{3-4}
                                &                                          & Data space optimization &~\cite{nigam2020augmenting, xie2020mars, khalifa2020distributional, fu2021differentiable,  kang2020gratis,kang2018efficient} \\\cline{2-4}
                                & \multirow{2}{*}{Peri-generation control} & Goal-oriented control &~\cite{dathathri2019plug, yu2017seqgan,you2018graph, hu2018deep, de2018molgan, guo2018long, putin2018reinforced, tambwekar2018controllable, shi2020graphaf, yang2020practical, samanta2020nevae, wang2021multi} \\\cline{3-4}
                                &                                          & Distribution-guided control & ~\cite{ gebauer2019symmetry,segler2018generating, yang2018aag}\\
    \hline
    \multirow{4}{*}{\thead{Controllable transformation\\ from source data}} & \multirow{2}{*}{Fixed transformation} & Within-domain transformation & ~\cite{prabhumoye2018style, jin2018learning, sood2018application, luo2021graphdf, fu2021differentiable, fu2020core, sudhakar2019transforming, hukkelaas2019deepprivacy, zhou2017optimizing}\\\cline{3-4}
                                    &                                             & Cross-domain transformation &~\cite{ tits2019exploring, liu2021reinforcement, sood2018application, jin2018learning, fu2020core} \\\cline{2-4}
                                    & \multirow{2}{*}{Steerable transformation} & Control via reference data &~\cite{hsu2018hierarchical, chen2019controllable, liang2019pcgan, chen2019hierarchical, maximov2020ciagan, li2021controllable, huang2021generating, pumarola2020c, yang2018unsupervised, valle2020mellotron,kurihara2021prosodic, kim2021expressive, liu2021reinforcement, bian2019multi,inoue2021model, cai2021emotion, krause2020gedi,song2018talking,kwon2019effective, tits2019methodology,  sini2020introducing,li2018delete,tits2019visualization} \\\cline{3-4}
                                    &                                            & Control via latent space &~\cite{fabbro2020speech, john2018disentangled, abdal2021styleflow, zhu2021low, xia2021gan, thermos2021controllable, chang2021changing, habib2019semi, shen2020interpreting, qiao2020sentiment, guu2018generating,ren2019fastspeech,luocontrollable,raitio2020controllable, ren2019fastspeech,cui2021emovie,engel2020ddsp} \\
    \hline

    \end{tabular}
    \end{adjustbox}
    \label{tab:data_summary}
    % \vspace{-4mm}
\end{table}

\subsection{Performance Evaluations}
\label{sec:eval}
The evaluation of the performance of controllable deep data generation consists of two parts: (1) \textit{data-quality evaluation} evaluates the general quality of generated data and (2) \textit{property-controllability evaluation} evaluates how well the properties are controlled. 

\textbf{Data-quality evaluation} of controllable deep data generation is adapted from the evaluation metrics for general deep generative models. The data-quality evaluation consists of two strategies: (1) self-quality-based metrics and (2) distribution-based metrics. Self-quality-based metrics are conducted to evaluate the overall quality of generated data including novelty, validity and uniqueness. Novelty measures how different the generated data is from the training data. For instance, in the task of graph generation, novelty measures the percentage of generated graphs that are not sub-graphs of the training graphs. Validity, as its name suggests, measures whether generated data preserves specific characteristics. For example, validity for molecule generation tasks measures the percentage of generated molecules that are chemically valid based on domain-specific rules~\citep{you2018graph}. Since ideally the generated data should be diverse but not identical, uniqueness measures the diversity of generated data calculated by the percentage of unique data in all generated data. Distribution-based metrics measure the distance between the distribution of generated data and the distribution of real data. Lower value of these metrics corresponds to similar distribution preserved by the generated and real data and indicates better generation results. A few metrics are designed to measure the distance between distributions, such as Kullback–Leibler divergence (KLD)~\citep{guo2021deep}, Maximum Mean Discrepancy(MMD)~\citep{sutherland2016generative, zhang2021tg} and Fr\'echet Inception Distance (FID)~\citep{song2019generative}. Since data-quality evaluation can be applied to any generation tasks and has been concluded in a few surveys~\citep{guo2020systematic, cheng2021molecular, harshvardhan2020comprehensive}, we will not regard it as the focus of our survey, and details can be referred to these surveys.

\textbf{Property-controllability evaluation} that assesses how well the properties are controlled is a critical but challenging task since generated data usually does not explicitly tell us if those properties are preserved. For instance, we can surely tell if an image contains a specific pattern by our eyes, but we cannot tell directly the toxicity of generated molecules in this way. In general, metrics for property-controllability evaluation are designed to measure the distance between properties of generated data and target properties, which comprise two strategies: (1) automated evaluation-based metrics and (2) human annotation-based metrics. The typical evaluation metrics in evaluating controllable deep data generation are summarized in Table~\ref{tab:eval}.

\begin{table}[]
\caption{Metrics for property-controllability evaluation.}
\begin{adjustbox}{width=\textwidth}
\begin{tabular}{|lll|l|l|}
\hline
\multicolumn{3}{|l|}{Type}                                                                                                                                                                           & Metrics            & Formula \\ \hline\hline
\multicolumn{1}{|l|}{\multirow{12}{*}{\thead{Automated\\evaluation}}} & \multicolumn{1}{l|}{\multirow{7}{*}{\thead{Continuous\\properties}}} & \multirow{4}{*}{Distance-based metrics}       & MSE  & $\frac{1}{M}\vert\vert c-c^*\vert\vert_2^2$\\ \cline{4-5} 
\multicolumn{1}{|l|}{}                                                     & \multicolumn{1}{l|}{}                                                   &                                               & MAE                &  $\frac{1}{M}\vert\vert c-c^*\vert\vert_1$       \\ \cline{4-5} 
\multicolumn{1}{|l|}{}                                                     & \multicolumn{1}{l|}{}                                                   &                                               & Cosine similarity  &  $\frac{c\cdot c^*}{\vert\vert c\vert\vert\cdot\vert\vert c^*\vert\vert}$\\\cline{4-5} 
\multicolumn{1}{|l|}{}                                                     & \multicolumn{1}{l|}{}                                                   &                                               & L2 norm            &  $\vert\vert c - c^*\vert\vert_2$       \\ \cline{3-5} 
\multicolumn{1}{|l|}{}                                                     & \multicolumn{1}{l|}{}                                                   & \multirow{2}{*}{Likelihood-based}     & CLL                &  $\log p(c\vert x)$       \\ \cline{4-5} 
\multicolumn{1}{|l|}{}                                                     & \multicolumn{1}{l|}{}                                                   &                                               & AMP                &  $AMP_i=\frac{1}{M}\sum_{m=1}^M p_m(x_i), i = 1,...,N$       \\ \cline{3-5} 
\multicolumn{1}{|l|}{}                                                     & \multicolumn{1}{l|}{}                                                   & Hypothesis testing-based              & p-value            &  $-$  \\ \cline{2-5} 
\multicolumn{1}{|l|}{}                                                     & \multicolumn{1}{l|}{\multirow{5}{*}{\thead{Discrete\\properties}}}   & \multirow{4}{*}{Classification-based} & Accuracy           &  $\frac{TP + TN}{M}$       \\ \cline{4-5} 
\multicolumn{1}{|l|}{}                                                     & \multicolumn{1}{l|}{}                                                   &                                               & Precision          &  $\frac{TP}{TP+FP}$ \\ \cline{4-5} 
\multicolumn{1}{|l|}{}                                                     & \multicolumn{1}{l|}{}                                                   &                                               & Recall          & $\frac{TP}{TP+FN}$ \\ \cline{4-5} 
\multicolumn{1}{|l|}{}                                                     & \multicolumn{1}{l|}{}                                                   &                                               & F1 score           & $2\times\frac{\text{precision}\times\text{recall}}{\text{precision}+\text{recall}}$ \\ \cline{3-5} 
\multicolumn{1}{|l|}{}                                                     & \multicolumn{1}{l|}{}                                                   & Entropy-based                        & BCE                &  $-\frac{1}{M}\sum_{m=1}^M \{c_i\log p(c_i)+(1-c_i)\log(1 - p(c_i))\}$  \\ \hline
\multicolumn{1}{|l|}{\multirow{6}{*}{\thead{Manual\\validation}}}      & \multicolumn{2}{l|}{\multirow{4}{*}{Score-based}}                                                               & MOS                &  $\frac{1}{S}\sum_{i=1}^{S}RATE_i$       \\ \cline{4-5} 
\multicolumn{1}{|l|}{}                                                     & \multicolumn{2}{l|}{}                                                                                                   & Accuracy           & $\frac{TP + TN}{M}$ \\ \cline{4-5} 
\multicolumn{1}{|l|}{}                                                     & \multicolumn{2}{l|}{}                                                                                                   & Precision          & $\frac{TP}{TP+FP}$ \\ \cline{4-5} 
\multicolumn{1}{|l|}{}                                                     & \multicolumn{2}{l|}{}                                                                                                   & Recall          & $\frac{TP}{TP+FN}$ \\ \cline{4-5} 
\multicolumn{1}{|l|}{}                                                     & \multicolumn{2}{l|}{}                                                                                                   & F1 score           & $2\times\frac{\text{precision}\times\text{recall}}{\text{precision}+\text{recall}}$        \\ \cline{2-5} 
\multicolumn{1}{|l|}{}                                                     & \multicolumn{2}{l|}{Rank-based}                                                                                 & Average human rank &    $\frac{1}{S}\sum_{i=1}^{S}RANK_{ij}$     \\ \hline
\end{tabular}
\end{adjustbox}
\label{tab:eval}
% \vspace{-2mm}
\end{table}

Automated evaluation-based metrics are distance measures formularized to compute the similarity of properties of the generated data and the target properties. Two steps are contained: property calculation and property validation, which can be broken down to the validation of continuous and discrete values. Specifically, the property values in this way are first extracted from the generated data either by directly calculation or via pre-trained models to be compared with the target properties~\citep{mirza2014conditional, odena2017conditional, hu2017toward, bodla2018semi, xu2019modeling, yang2021cpcgan}. Then, for continuous properties, automated evaluation-based metrics include distance measures (e.g., mean squared error (MSE), mean absolute error (MAE), cosine similarity, and L2 norm)~\citep{camino2018generating, moon2020conditional, zhou2019talking, mirza2014conditional, john2018disentangled}, likelihood-based metrics (e.g., conditional log-likelihoods (CLL) and Attribute Mean Probability (AMP)) if properties are assumed to follow a specific probabilistic distribution~\citep{sohn2015learning, park2018data, kingma2018glow, liu2019conditional} and hypothesis testing-based metrics (e.g., p-values from A/B testing) to test if properties of generated data and real data are significantly different~\citep{henter2018deep, yang2018aag, esteban2017real} (Table~\ref{tab:eval}). Note that to calculate AMP, $M$ different classifiers are trained for $M$ different properties in $c$. Then for each data point $x_i$ in $x$, the classifiers could output the probabilities $p_m(x_i)$ as presented in Table~\ref{tab:eval}. For discrete properties, automated evaluation-based metrics include classification-based metrics (e.g., accuracy, precision and F1 score)~\citep{liu2021reinforcement, prabhumoye2018style, cui2022can} and entropy-based metrics (e.g,. binary cross entropy (BCE))~\citep{hu2018deep, goetschalckx2019ganalyze}. Classification-based metrics are computed based on true positive (\textit{TP}), true negative (\textit{TN}), false positive (\textit{FP}) and false negative (\textit{FN}) when comparing predicted properties and true properties. Specifically, Accuracy is calculated as the percentage of properties from the generated data that are in the same class as the real properties. F1 score views the generator as a property classifier and measures how accurately the properties of generated data can be classified into the same class of the real properties~\citep{yacouby2020probabilistic}. BCE compares each of the predicted probabilities for properties of generated data to the real binary properties.

Human annotation-based metrics are obtained by recruiting human evaluators to score manually the properties of generated data~\citep{bodla2018semi, henter2018deep, shao2021controllable, tambwekar2019controllable, kedzie2020controllable, shoshan2021gan}. Strategies to obtain human annotated score can be classified into two types: (1) score-based metrics and (2) rank-based metrics. Score-based metrics directly measure the quality of the properties of generated data. For instance, some works directly score the quality of generated speech such as the mean opinion score (MOS), in which $S$ evaluators are asked to rate the quality and naturalness of the stimulus $i$ (i.e., $RATE_i$) on a Likert-type scale from 1 to 5 with scores labeled as Bad, Poor, Fair, Good and Excellent, respectively~\citep{rosenberg2017bias}. Other works recruit human evaluators to classify generated data according to their properties and calculate the classification-based score such as accuracy, precision, and F1 score compared to real properties~\citep{guu2018generating, sudhakar2019transforming}. Rank-based metrics compare performance of controllable deep data generation across models. Human evaluators are asked to rank subjectively the data generated by various models according to the quality of properties of generated data. Then the average human rank is calculated for each model with $RANK_{ij}$ as the rank of the evaluator $i$ on the model $j$. The model with the lower average human rank should have better performance.

\section{Techniques for Controllable Deep Data Generation}
\label{sec:tech}
In this section, we introduce the techniques of controllable generation from scratch and the controllable transformation from source data in detail in Secton~\ref{sec:gen} and Section~\ref{sec:trans}, respectively.

\subsection{Controllable Generation from Scratch}
\label{sec:gen}
% Data generation aims to learn the distribution $p_{G}(x)$ from the observed data sampled from the real distribution $p(x)$ \citep{du2021graphgt}. In this process, controllable generation from scratch aims to generate data conditioning on the target properties $c=c^*$ from $x\sim p_{G}(x\vert c=c^*)$. 
According to which phase of the generation process the control is involved, \emph{controllable generation from scratch} can be divided into three scenarios: (1) \emph{pre-generation control} feeds the property-control signal to the model before the data is generated; (2) \emph{post-generation control} iteratively optimizes the properties of the data after the data is generated; (3) \emph{peri-generation control} directly generates data with desired properties by sampling from the data distribution learned during the training process of the model. 

These three tactics have their specific advantages and limitations. Pre-generation control simply manipulates the value of properties as the input of the generator before the generation process and is relatively easy to implement. Besides, the generator takes the property-related information along and disentangles it with the data representation as the input, preserving a high level of interpretability in the model. Pre-generation control cannot guarantee that the target properties are preserved by the generated data, however, since it does not evaluate and leverage the properties of generated data in the generation process. By contrast, post-generation control evaluates the properties after the data is generated and leverages the properties to guide the generation process for the next iteration, until the desired objective of properties is reached. Hence, the post-generation control usually guarantees that generated data preserves desired properties. Nevertheless, post-generation control requires that the properties are evaluated after the data is generated and serve as the control signal to improve the generation process. This usually leads to a more complex generation architecture compared to pre-generation control-based techniques. Peri-generation control directly generates data via the learned distribution, which owns a relatively simple architecture of generation. Peri-generation control usually requires a relatively large dataset to train the model, however, it entails that the model should learn the distribution of the data with desired properties during the training process. 

\begin{figure}[h]
\begin{center}
\includegraphics[width=0.85\textwidth]{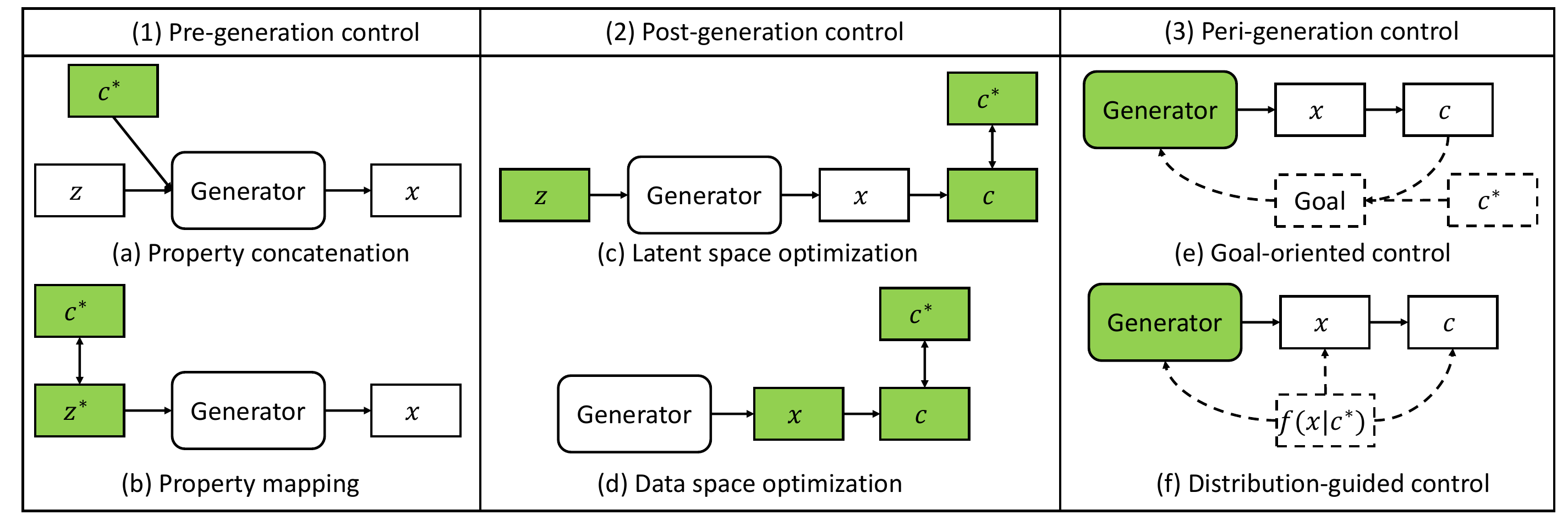}
% \vspace{-2mm}
\caption{Architecture of control via controllable generation from scratch. Green modules are where the control signal is entailed.}
\label{fig:generation}
% \vspace{-2mm}
\end{center}
\end{figure}

\subsubsection{Pre-generation Control}\hfill\\
Pre-generation control, as shown in Fig. \ref{fig:taxonomy} and Fig.~\ref{fig:generation} (1), aims to control properties of generated data conditioning on target properties. In general, two different ways are available for incorporating the control signal of desired properties as the input: (a) \textit{property concatenation} feeds the raw properties to the generator while (b) \textit{property mapping} first encodes the properties into their embedding and feeds the embedding to the generator.

Compared with property concatenation techniques, the structure of property mapping-based models is usually more complex, since the embedding of the target properties is obtained by either encoding the target properties into the latent space via an encoder, or optimizing the latent space, based on the learned mapping between the latent space and properties. Nevertheless, different types of input properties (i.e., continuous or categorical properties) may cause difficulties in the design of the generator of property concatenation-based techniques, which is instead not a challenge for property mapping-based techniques since they only take continuous property embedding as the input.

\begin{figure}[h]
\begin{center}
\includegraphics[width=0.8\textwidth]{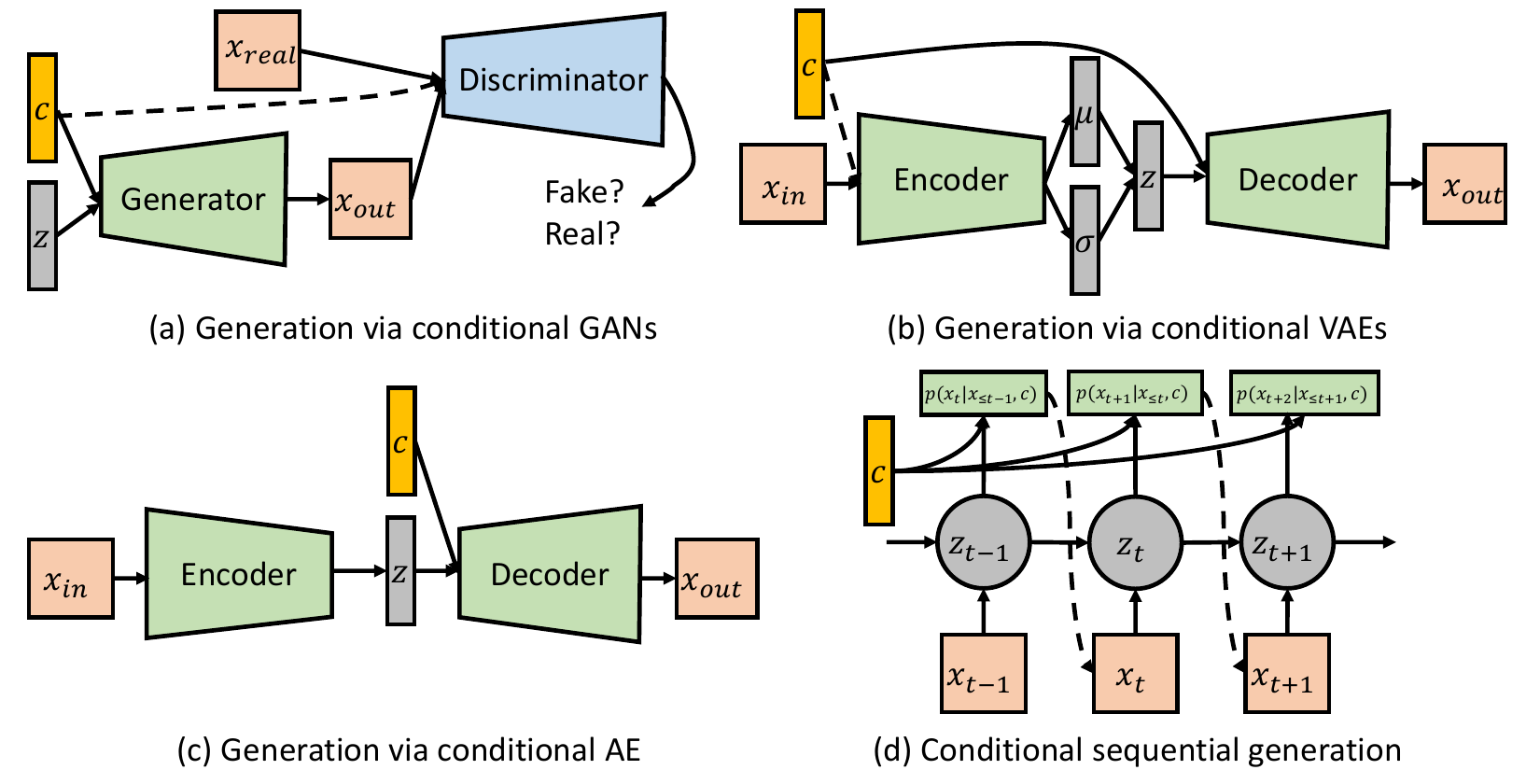}
% \vspace{-2mm}
\caption{Property concatenation-based models for controllable generation from scratch.}
\label{fig:prop_concat}
\end{center}
% \vspace{-4mm}
\end{figure}

\textbf{Property concatenation} conditions the generation process on the original target properties simply by feeding property codes along with the data representation into the generator, as shown in Fig.~\ref{fig:generation} (a) and Fig. \ref{fig:prop_concat}. As a result, given the target properties $c^*$, the generation process can be formulated as $x\sim p(x\vert c=c^*)$. In this process, the properties are independent of the latent representation of the data. 

In recent years, a huge amount of effort has been devoted to controllable generation from scratch with the strategy of property concatenation. Techniques to achieve property concatenation have been developed under popular existing deep generation frameworks, such as adversarial training, variational inference, autoencoder and sequential model. For instance, the framework of adversarial training has been borrowed by works that control data generation by conditioning on the auxiliary information \citep{mirza2014conditional}. As presented in Fig. \ref{fig:prop_concat} (a), several works feed handcrafted properties into the generator of GANs to realize property control on generated data \citep{hu2017toward, prabhumoye2020exploring, xu2019modeling}, which have the objective as follows:
\begin{align}
    \mathcal{L} = \mathbb{E}_{x\sim p(x)}[\log D(x)] + \mathbb{E}_{z\sim p(z)}[\log (1-D(G(z\vert c)))],
    \label{eq:condgan_gd}
\end{align}
where $D$ and $G$ are discriminator and generator, respectively. Based on Eq.~\ref{eq:condgan_gd}, $D$ and $G$ are trained together, in which $G$ is trained conditional on desired properties to produce realistic data by having it compete against $D$ in a game-theoretic framework. In addition to feeding properties into the generator, some works also input the properties into the discriminator \citep{hu2017toward, prabhumoye2020exploring, xu2019modeling}. In this case the $D(x)$ of the Eq. \ref{eq:condgan_gd} can be rewritten as $D(x\vert c)$ to condition the discrimination process on the properties.

In addition to training on the adversarial loss, another way to train the objective is to maximize the joint likelihood of data conditional on the target properties based on the variational inference. The variational inference is proposed to approximate the intractable posterior density of the latent variables for example, approximate the posterior $p(z\vert x, c)$ when training conditional VAEs that are broadly used for controllable generation from scratch-based frameworks. For implementation, several works feed the properties into the decoder of VAEs along with the data representation (Fig. \ref{fig:prop_concat} (b)) \citep{kingma2014semi, henter2018deep, tan2020generative}. The objective can be formularized as follows:
\begin{align}
    \mathcal{L} = -\mathbb{E}_{z\sim p(z\vert x)}[\log p_{\theta}(x\vert z, c)]+KL[q_{\phi}(z\vert x)\vert\vert p(z)],
    \label{eq:condvae_d}
\end{align}
where $q_{\phi}(z\vert x)$ is the approximated posterior distribution of $p(z\vert x)$ and $\phi$ is the parameter of the encoder. $p_{\theta}(x\vert z, c)$ is the decoder that generates data from latents parameterized by $\theta$, where properties $c$ are concatenated with the latent representation of data $z$. Under the same framework, other works feed properties both into the encoder with data and into the decoder with the data representation \citep{henter2018deep, jin2020multi}. For example, given target properties, RationaleRL encodes rational of properties and data into latent space by the encoder. The decoder is initialized with both rational and latent data representation to ensure that generated molecules contain the corresponding rational. In this case, $p(z\vert x)$ in the objective above becomes $p(z\vert x, c)$ and the approximated posterior is changed to $q_{\phi}(z\vert x, c)$.

Besides techniques under the framework of adversarial training and variational inference, autoencoder serves as another framework for controllable generation from scratch (Fig.~\ref{fig:prop_concat} (c)). Similar to the variational inference-based techniques mentioned above, several works directly feed the target properties into the decoder \citep{cheung2014discovering} of autoencoder to generate data with desired properties. The objective can be formulated as the reconstruction loss:
\begin{align}
    \mathcal{L} = -\mathbb{E}_{z\sim p(z\vert x)}[\log p_\theta(x\vert z, c)],
    \label{eq:condae_d}
\end{align}
where $p_\theta(x\vert z, c)$ is the generator that is learned by optimizing Eq.~\ref{eq:condae_d} and the concatenation of data representation and properties severve as the input. Another branch of property concatenation-based techniques is implemented by sequential generative models, particularly for generating sequential data. Sequential generative models condition the generation of the token at the current stage on both previously generated tokens and the target properties (Fig. \ref{fig:prop_concat} (d)) \citep{ficler2017controlling, keskar2019ctrl, li2018learning, xu2020megatron}:
\begin{align}
    x_t\sim p_\theta(x_t\vert x_{\le t-1}, c),
    \label{eq:condseq}
\end{align}
where $x_t$ is generated at the $t$-th stage and $x_{\le t-1}$ is the set of data generated before the $t$-th stage. The generator is parameterized by $\theta$ and properties are concatenated with data generated at the previous step. The condition sequential generation has a broad range of applications. For instance, several deep text generative models view a sentence as a sequence of words and generate the whole sentence word-by-word sequentially \citep{ficler2017controlling, keskar2019ctrl, xu2020megatron}. To realize the controllable data generation, the word to be generated at the current stage depends on both previously generated words and the target properties. Controllable graph generation can also be conducted in a sequential manner, where the node to be generated at the current stage relies on both previously generated nodes, edges, and the target properties \citep{li2018learning}. In addition to the autoregressive model, generative diffusion is a rapidly emerging method for controllable data generation. Most strategies based on property concatenation rely on classifier-free diffusion guidance~\citep{ho2022classifier}. This guidance blends score estimates from both a conditional diffusion model and a concurrently trained unconditional diffusion model. Specifically, rather than learning $\epsilon_\theta(x_t, t)$, the classifier-free diffusion guidance learns $\epsilon_\theta(x_t, c, t)$ in which the property is concatenated with latents of the data (i.e., diffused data). Hoogeboom et al.~\cite{hoogeboom2022equivariant} generates molecules conditional on their properties by concatenating the property values with the latents. Tevet et al.~\citep{tevet2022human} generates human motion based on text descriptions similarly following the property-concatenation strategy. Nichol et al.~\citep{nichol2021glide} proposed GLIDE, which serves as a text-conditional diffusion model that has a superior performance by means of classifier-free guidance training strategy. Huang et al.~\citep{huang2022fastdiff} generate speech based on the diffusion model conditioning on the Mel-spectrogram at each step.

\begin{figure}[h]
\begin{center}
\includegraphics[width=0.8\textwidth]{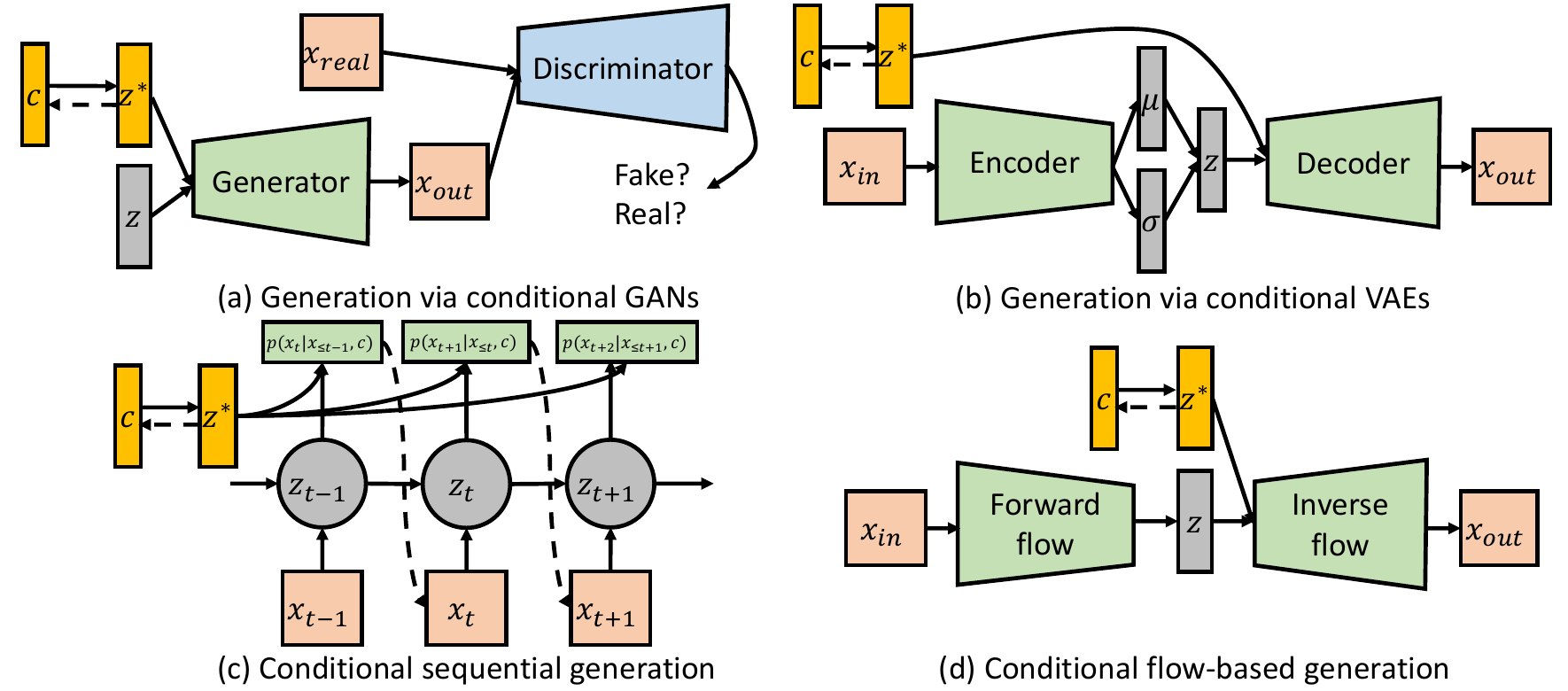}
\caption{Property mapping-based models for controllable generation from scratch.}
\label{fig:prop_map}
% \vspace{-3mm}
\end{center}
\end{figure}

\textbf{Property mapping} first maps the target properties into latent space, as shown in Fig.~\ref{fig:generation} (b) and Fig. \ref{fig:prop_map}, and then enables property control by feeding the embedding of the target properties into the generator. Suppose the mapping from the target properties to the corresponding latent variables is achieved by the function $f(c)$ and represented by $z^*=f(c^*)$, then given the target properties $c^*$, generated process can be formulated as $x\sim p(x\vert z^*=f(c^*))$.

Similar to property concatenation, property mapping techniques designed for controllable generation from scratch are mostly under the framework of adversarial training \citep{moon2020conditional, shoshan2021gan}, maximizing joint likelihood via the variational inference \citep{jin2018junction, sohn2015learning, shao2021controllable, guo2020property, chen2020music} and conditional sequential generation \citep{vasquez2019melnet}. As illustrated in Fig. \ref{fig:prop_map} (a), unlike the adversarial training-based models for property concatenation that can feed the target properties into both generator and discriminator, the conditional GANs for property mapping only feed the embedding of target properties into the generator since the discriminator of GANs aims to classify fake and real data rather than their embedding \citep{moon2020conditional, shoshan2021gan}. The objective function of conditional GANs can be adapted from Eq. \ref{eq:condgan_gd}:
\begin{align}
    \mathcal{L} = \mathbb{E}_{x\sim p(x)}[\log D(x)] + \mathbb{E}_{z\sim p(z)}[\log (1-D(G(z, z^*\vert c)))],
    \label{eq:condgan_pm}
\end{align}
where $z^*$ is the embedding of the input properties obtained from $z^*=f(c)$. Models learned based on the variational inference act in a similar way as in the property concatenation setting. As shown in Fig. \ref{fig:prop_map} (b), the generation is controlled by feeding the embedding of the target properties into the decoder \citep{jin2018junction, sohn2015learning, shao2021controllable, guo2020property, chen2020music}. Specifically, the objective can be formularized as:
\begin{align}
    \mathcal{L} = -\mathbb{E}_{z^*\sim p(z^*\vert c),z\sim p(z\vert x)}[\log p_\theta(x\vert z, z^*)]+KL[q_{\phi_1}(z\vert x)\vert\vert p(z)] + KL[q_{\phi_2}(z^*\vert c)\vert\vert p(z^*)],
    \label{eq:condvae_pm}
\end{align}
where $\phi_1$ and $\phi_2$ are parameters of posteriors to approximate $p(z)$ and $p(z^*)$, respectively. The decoder is parameterized by $\theta$. The existing works have shown considerable diversity in embedding properties into latent space. For instance, some works pre-train a property encoder to extract and embed the target properties into the latent space. Then the data can be generated by the decoder that takes both the data and property embedding as the input \citep{shao2021controllable, chen2020music}. In some other works, the mutual dependence between the individual property and the specific latent variable can be realized via an invertible function \citep{guo2020property}. Then in the generation process, latent variables can be backtracked given target properties via the invertible function and serve as the input of the decoder. In addition, prediction models such as sparse Gaussian process can also be used to predict properties from the data representation \citep{jin2018junction}. The corresponding latent variables can be optimized based on techniques like Bayesian optimization~\citep{griffiths2020constrained} given target properties and are then fed into the decoder to generate data. Particularly, for sequential data, as presented in Fig. \ref{fig:prop_map} (c), conditional information of properties are projected onto the input layer along with the data from the previous stage to generate data at the current stage: $x_t\sim p(x_t\vert x_{\le t-1}, z^*=f(c))$ \citep{vasquez2019melnet}. In addition to generating sequential data, Choi et al.~\citep{choi2021ilvr} employ the diffusion model to generate images while at each step the generation is conditioned on the control signal extracted from the reference image. 
% \begin{align}
%     x_t\sim p(x_t\vert x_{\le t-1}, z^*=f(c)).
%     \label{eq:condseq_pm}
% \end{align}

In addition to the models described above, as shown in Fig.~\ref{fig:prop_map} (d), conditional flow-based models relying on the invertible constraint serve as another tool for property mapping-based controllable generation from scratch by learning the mapping between data distribution $p(x)$ and prior distribution $p(z)$ using a bijective function $z=F(x)$ \citep{zang2020moflow, liu2019conditional, lu2020structured, ardizzone2019guided}. Then the data is generated by the invertible mapping from the latent variables. One type of conditional flow-based generative model encodes properties into the latent space $z^*$ via the encoder, and encodes the data into $z$ via a forward flow \citep{liu2019conditional}. The distance of the distribution of the property embedding and data embedding is minimized. Then the data can be generated from the embedding of the target properties. To train this conditional flow-based generative model, we maximize the likelihood:
% \vspace{-5mm}
\begin{equation}
    \log p_\theta(x, c) = \log p_\theta(z, c) + \log \vert det\frac{\partial F(x)}{\partial x} \vert.
    \label{eq:flow}
    % \vspace{-6mm}
\end{equation}
Another type of conditional flow-based generative model feeds properties as the parameter of forward function, which is usually a neural network \citep{lu2020structured, zang2020moflow, ardizzone2019guided}. During the generation process, the target properties are fed into the inverse function to generate corresponding data. In the training process, we modify $p_\theta(x, c)$ to $p_\theta(x\vert c)$ and $F(x)$ to $F(x;c)$ in objective Eq.~\ref{eq:flow} while $z = F(x;c)$.

\subsubsection{Post-generation Control}\hfill\\
Post-generation control, as suggested in Fig.~\ref{fig:generation} (2), achieves controllable generation from scratch by first generating data and then optimizing the properties of generated data towards the target property $c^*$. According to whether latent space is entailed during the process of controllable generation from scratch, two common tactics have been employed: (a) \textit{latent space optimization} takes advantage of the continuous latent vectors learned by models to avoid the combinatorial search of data structures. Specifically, it manipulates the learned latent vectors by optimization methods or heuristics; by contrast, (b) \textit{data space optimization} does not embed data or properties into latent space, but rather directly generates data with target properties by searching over the original data space which is often high-dimensional and sometimes discrete. 

The key difference between latent space optimization and data space optimization is whether a latent space is entailed during the process of controllable generation from scratch. The benefit of latent space optimization is that the latent space learned by models is usually continuous and low-dimensional which makes the optimization phase an easier problem. In addition, regularization or structure prior can be easily imposed while learning the latent space, which is particularly true when the input data is highly discrete, high-dimensional and indifferentiable such as graphs, texts, etc. By contrast, data space optimization usually optimizes the property-related objective based on the domain knowledge The objective function is usually easier to optimize than the optimization on the latent space which is even non-convex.

\textbf{Latent space optimization} often requires a latent space learned by models which have an encoder $q_{\phi}(z|x)$ that maps data from data space $X$ to the learned latent space $Z$, and a decoder that maps the latent vector back to the data space. The encoder and decoder usually are parameterized by neural networks depending on the data type (e.g., GNN for graph data, CNN for image data, etc.). As shown in Fig.~\ref{fig:generation} (c), Latent space optimization refers to optimizing the latent vectors $z$ by a heuristic function $f(z)$ in the latent space to obtain optimized latent vectors $z'$ that can be decoded to $x'$ with desired properties, as follows:
\begin{align}
z = q_{\phi}(z|x); \quad z' = z + f(z); \quad x' = p_{\theta}(x'|z'),
\end{align}
where $f(z)$ can be in different formats and learned in different settings.

The first way to learn $f(z)$ is to find meaningful latent directions aligned with the change of properties, then controllable generation from scratch can be easily achieved by moving latent vectors along the latent directions. Some works~\citep{upchurch2017deep,kingma2018glow,madhawa2019graphnvp} simply interpolate linear direction between the data with and without desired properties as the directions to control the properties while others ~\cite{goetschalckx2019ganalyze,jahanian2019steerability,plumerault2020controlling} also discover linear directions that control the properties but in an end-to-end manner guided by a discriminator that predicts the properties. H{\"a}rk{\"o}nen et al. ~\cite{harkonen2020ganspace} leverages PCA to decompose the latent space learned by the model to factorize controllable latent directions. Shen and Zhou~\cite{shen2021closed} discover latent directions that control semantically meaningful properties of data by decomposing generator parameters without the need to access semantic labels as in previous methods. Dhasarathy et al.~\cite{parthasarathy2020controlled} leverages a GANs-based model and linearly interpolates in a metrics space which correlates with the properties instead of directly in the latent space. 

The second way to find $f(z)$ is to leverage regularization over the latent space and manipulate the latent vectors to achieve controllable generation from scratch by following the bias introduced to the model. Johnson et al.~\cite{johnson2016composing} propose a new framework that learns a structured latent-variable model for controllable latent variables in a structured representation. Ma et al.~\cite{ma2018constrained} aim to generate semantically valid graphs and formulates semantic constraints as regularization terms to generative models. Guo et al.~\citep{guo2021deep,guo2020interpretable}, Yang et al.~\citep{yang2020dsm}, Li et al.~\citep{li2021editvae} and Zhou et al.~\cite{zhou2019talking} introduce inductive biases over the structure of the studied data and propose a structured latent-variable generative model that learns controllable latent variables in an unsupervised learning manner. Pati and Lerch~\citep{pati2021disentanglement}, Du et al.~\citep{du2020interpretable} and Xu et al.~\cite{xu2020variational} leverage a disentangled VAEs model that regularizes the latent variables for controlling properties of data. 

The third way to find $f(z)$ is the optimization-based method, in which a heuristic function is learned to estimate properties $c$ of a given $z$. With the learned heuristic functions, optimization-based methods can be applied to search $z'$ with desired properties. For instance, Jin et al.~\citep{jin2018junction} and Zang and Wang~\cite{zang2020moflow} optimize the latent vectors learned by the generative models for generating data with desired properties.

\textbf{Data space optimization} directly optimizes over the data space to obtain data $x'$ with desired properties instead of learning a continuous latent space, as illustrated in Fig.~\ref{fig:generation} (d), unlike latent space optimization. Even though data space optimization suffers from the high-dimensional and usually discrete searching problem, it does not require the mapping to continuous space which sometimes causes loss of information: $x' = f(x)$.

Kang et al.~\citep{kang2018efficient, kang2020gratis} and Nigam et al.~\citep{nigam2020augmenting} design genetic algorithm with a neural network (often as a property evaluator) to search data with desired properties. Xie et al.~\citep{xie2020mars} devise an MCMC-based approach for controllable generation from scratch that samples data maximizing the distribution of desired properties. Fu et al.~\citep{fu2021differentiable} formulate data in a differentiable format which allows direct optimization over the data space. Song et al.~\citep{song2020score} borrows a score-based generative diffusion model to generate data conditioning on the target properties. The intuition is that at the last step of the reverse diffusion process, the data is sampled from $p_0(x_0\vert c^*)$. This only requires to compute $p_t(c^*\vert x_t)$ at all steps that can be modeled as the property predictor using neural networks. Additionally, classifier-guidance generative diffusion is another approach that guides the model to generate data with desired properties by training a classifier and supervising the model with its gradient~\citep{dhariwal2021diffusion}. For instance, Kim et al.~\cite{kim2022guided} proposed Guided-TTS that achieved text-to-speech generation while controlling the target speaker. For classifier guidance, Guided-TTS utilizes an unconditional diffusion model trained on untranscribed speech, combined with a phoneme classifier from a large-scale speech recognition dataset. Kawar et al.~\citep{kawar2022enhancing} incorporated the time-dependent classifier's gradient into the diffusion process to encourages the generated image to be recognized as the target class. Note that in general classifier-guidance generative diffusion is not efficient as classifier-free diffusion guidance as it needs to compute gradient at each time step.

\subsubsection{Peri-generation Control}\hfill\\
Peri-generation control, as shown in Fig.~\ref{fig:generation} (3), learns the distribution of data and its relation to desired properties $c^*$ by supervised learning during the training process. In general, the distribution of data can be learned towards target properties in two ways: (a) \textit{goal-oriented control} initializes the generator with either random noise or an initialized sample, and optimizes the goal calculated by the target properties of generated data in an iterative manner; (b) \textit{distribution-guided control} trains the generator learned from the data with the desired properties. As a result, the distribution of data with desired properties is preserved by the generator during the training process. Since distribution-guided control generates data from the distribution by the observations, it usually cannot generate out-of-sample data as goal-oriented control does. 

Since the control signal is provided in the training process, moreover, distribution-guided control usually requires a more complicated training strategy such as transfer learning~\citep{zhuang2020comprehensive}. 

\textbf{Goal-oriented control} optimizes the goal regarding target properties and data, as shown in Fig.~\ref{fig:generation} (e), which is either designed based on domain knowledge~\citep{liu2021graphebm} or estimated from observations~\citep{arulkumaran2017deep}. Define $R(x, c)$ as the reward function, where $x$ is the data and $c$ is corresponding properties, then the goal-oriented control can be formularized as:
% \vspace{-1mm}
\begin{equation}
x =  \argmax_{x} R(x, c); \quad s.t.\quad c = c^*,
\label{eq:rwd}
% \vspace{-3mm}
\end{equation}
where $c^*$ is the target properties supposed to be preserved by the generated data. In implementation, $R(x, c)$ can be in different forms depending on the application to various domains and different generation techniques. For instance, Liu et al.~\cite{liu2021graphebm} devise an energy-based model that regards the reward function as the negative energy of molecules that incorporates the properties as part of the energy terms. Dathathri et al.~\cite{dathathri2019plug} takes the likelihood of the desired attribute as the reward function and introduces a plug-and-play method on a pre-trained language model that trains an additional property classifier to guide the generation process. Yang et al.~\cite{yang2020practical} deploys a distributed MCTS to the molecule generation problem that searches for molecules with desired properties where the reward score is provided by a chemistry simulator that evaluates the properties of the generated molecules. Samanta et al.~\cite{samanta2020nevae} manages to learn a property-optimized decoder for the VAEs-based model with a gradient-based algorithm to generate molecules with desired/optimal properties. In addition, the reward function in Eq.~\ref{eq:rwd} can also serve as the reward optimized by reinforcement learning-based techniques. For example, You et al.~\cite{you2018graph} utilizes a reinforcement learning approach to formulate the molecule generation problem which optimizes domain-specific rewards via policy gradient. Hu et al.~\cite{hu2018deep} expands the posterior regularization approach~\cite{ganchev2010posterior} that imposes knowledge constraints on posterior distributions of probabilistic models with the inverse reinforcement learning idea to allow the model to learn the constraints jointly with optimizing the objective. De Cao and Kipf~\cite{de2018molgan} formulate a GANs-based model under a reinforcement learning framework that utilizes the discriminator to guide the process to generate molecules with desired properties. Guo et al.~\cite{guo2018long} also formulates a GANs model under the reinforcement learning framework but additionally incorporates the high-level features learned by the discriminator into the generator to allow the control signal to be smooth over the long text rather than a single scalar value. Sanchez-Lengeling et al.~\citep{sanchez2017optimizing}, Putin et al.~\citep{putin2018reinforced} and De Cao and Kipf~\cite{de2018molgan} formulate the molecule generation problem under the reinforcement learning framework with a GANs model which utilizes the reward or discriminator network to provide feedback to generate molecules with desired properties. Tambwekar et al.~\cite{tambwekar2018controllable} formulates text generation under the reinforcement learning framework and introduces dense/intermediate reward rather than sparse reward at the end. Shi et al.~\cite{shi2020graphaf} focuses on the problem of molecule generation with a reinforcement learning framework that provides rewards for desired properties. 

\textbf{Distribution-guided control} lies in the fact that data generation models learn the distribution of the data $p(x)$ and bias the input data distribution towards the target distribution $p(x\vert c^*)$ of desired properties (i.e., $f(x\vert c^*)$ in Fig.~\ref{fig:generation} (f)). Thus, the model can generate data under the target distribution $p(x\vert c)$ with desired properties. For instance, Gebauer et al.~\cite{gebauer2019symmetry} employs an auto-regressive model for 3D molecule generation and biases the distributions of the target properties for controllable generation from scratch. To be specific, this model pre-selects 3D molecules with the target properties in desired ranges and fine-tunes the generative model with the subset of the whole training set to generate molecules with the target properties similar to the selected ones.  Segler et al.~\cite{segler2018generating} first pre-trains an auto-regressive generative model with a large dataset and then fine-tunes the model with a small subset of molecules known to be active against given biological targets. Yang et al.~\cite{yang2018aag} focus on the problem to generate high-quality audios embedding secret bits stream. Specifically, it proposes an auto-regressive model to generate audio based on the encoded bits stream as conditions.

\subsection{Controllable Transformation from Source Data}
\label{sec:trans}
\begin{figure}[h]
\begin{center}
\includegraphics[width=0.7\textwidth]{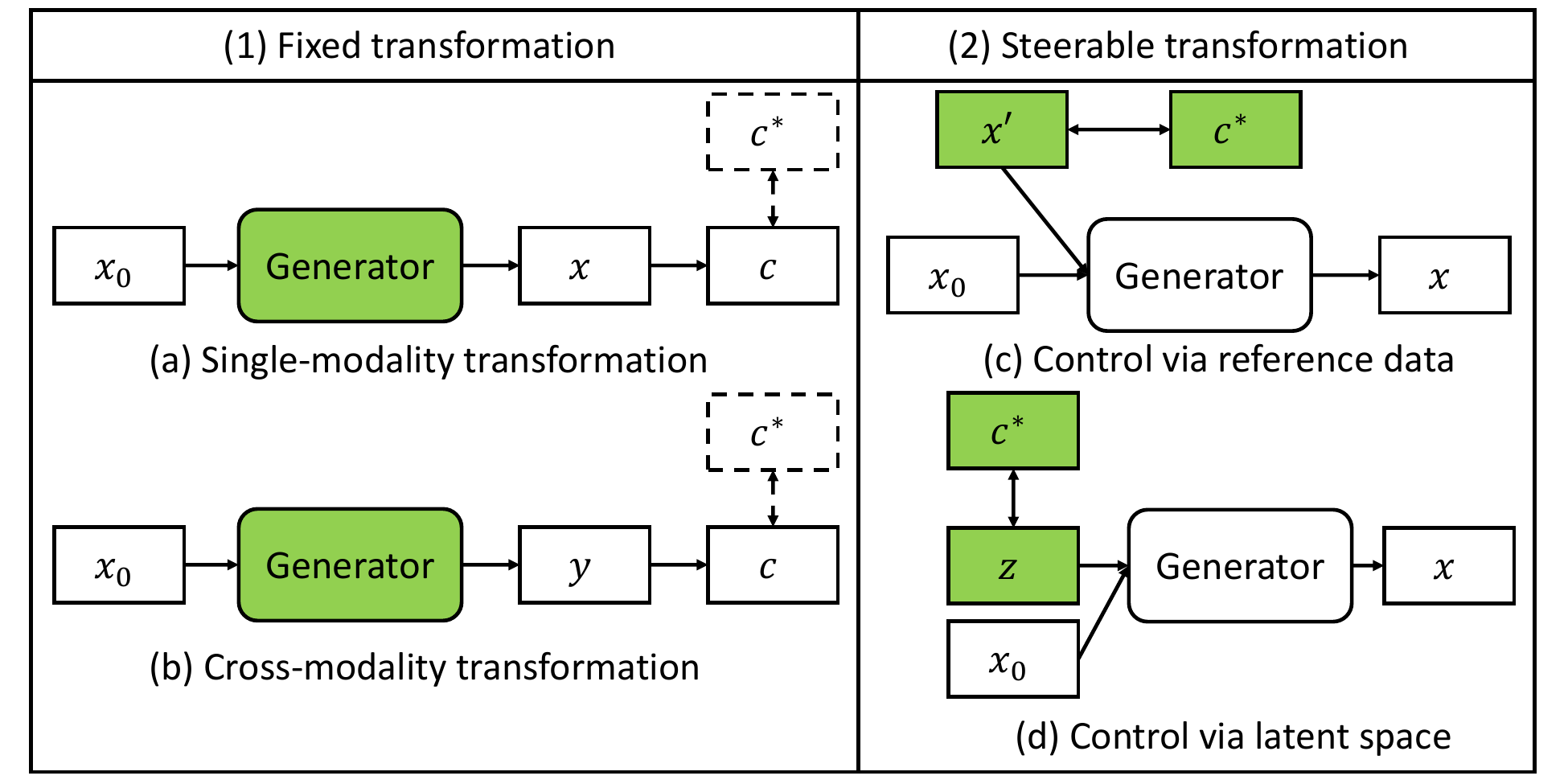}
\caption{Architecture of control via controllable transformation from source data. Green modules are the location of the control signal.}
\label{fig:transformation}
% \vspace{-3mm}
\end{center}

\end{figure}

Depending on whether the desired values of the properties are prefixed or steerable by the users, as presented in Fig.~\ref{fig:transformation}, controllable transformation from source data can be classified into two categories: (1) \textit{fixed transformation} only allows the source data to be transformed into the data with predefined value (or value change) of properties that are prefixed during model training. For instance, one may train an image translator for daytime-to-nighttime transformation, so that it can only transfer daytime scenery into its nighttime view. By contrast, (2) \textit{steerable transformation} enables users to input desired values (or value changes) of properties for guiding data transformation. For example, using steerable scenery-image transformation models, the model may take time (e.g., 1:25 pm) as the input from the user and then generate the scenery at that specified time based on the source scenery image. 

Steerable transformation may stay more interpretable, given that the control signal is explicitly fed into the model along with the source data during the generation process. By contrast, the fixed transformation handles the control signal in an implicit way, in that it takes only the source data as the input and instead learns the control signal during the training process. Nevertheless, the more flexible controlling ability is normally achieved at the cost of a more complex model structure to enable the properties to interact with the generator.
% \vspace{-3mm}

\subsubsection{Fixed Transformation}\hfill\\
Approaches for fixed transformation, as shown in Fig.~\ref{fig:transformation} (1), can be divided into two categories based on domains of the source data and the target data: (a) \textit{single-modality transformation} conducts property-controllable data transformation within the same modality, such as image-to-image, text-to-text, graph-to-graph, etc.; and (b) \textit{cross-modality transformation} translates the data from source modality to the data in another modality while controlling its properties. For example, one may transform text to audio while controlling the emotion expressed in the generated audio.
% \vspace{-3mm}
\begin{figure}[h]
\begin{center}
\includegraphics[width=0.9\textwidth]{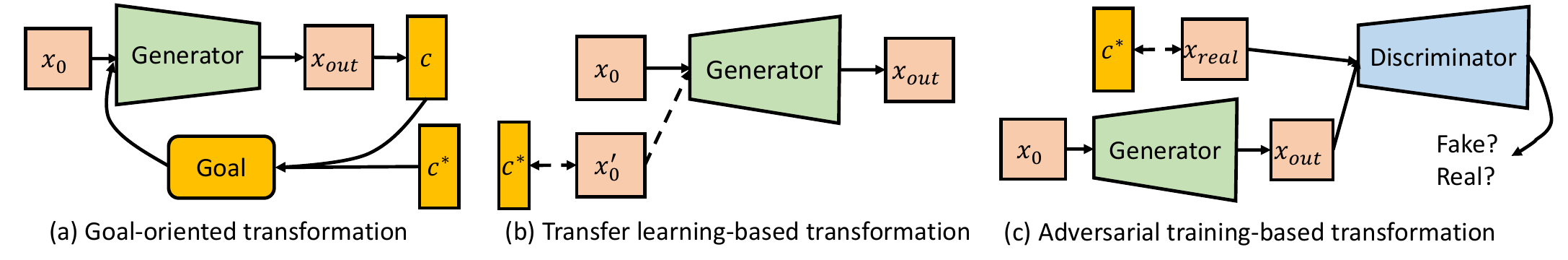}
% \vspace{-2mm}
\caption{Single-modality transformation for controllable transformation from source data.}
\label{fig:within_domain}
% \vspace{-5mm}
\end{center}

\end{figure}

\textbf{Single-modality transformation} instead translates the source data to target data in the same modality while changing properties of generated data in a prefixed way, as shown in Fig. \ref{fig:transformation} (a) and Fig. \ref{fig:within_domain}. For example, one may transform an image of human pose into another one with a predefined pose. As shown in Fig. \ref{fig:within_domain}, Single-modality transformation includes three types of techniques: 1) goal-oriented transformation; 2) transfer learning-based transformation and 3) adversarial training-based transformation. As shown in Fig. \ref{fig:within_domain} (a), goal-oriented transformation usually borrows the framework of reinforcement learning to generate data guided by the reward constructed via target properties \citep{luo2021graphdf, zhou2017optimizing, fu2021differentiable} with the same objective as Eq.~\ref{eq:rwd_trans}. For instance, Deep Reaction Optimizer optimizes certain chemical reactions by combining reinforcement learning with the domain knowledge of chemistry \citep{zhou2017optimizing}. GraphDF was designed to enforce optimized molecules to have desired properties by fine-tuning the model to optimize the score of properties under a reinforcement learning framework \citep{luo2021graphdf}. In addition to optimizing the reward with the reinforcement learning-based approach, other goal-oriented methods also have been explored for various tasks. The differentiable scaffolding tree controls the properties of generated data by optimizing the composite objective under a multi-objective optimization framework \citep{fu2021differentiable}: $\log \hat{x}^{(t+1)}  = \argmax_{x\in\mathcal{N}(\hat{x}^{(t)})}F(x)$, where $\hat{x}^{(t)}$ is the generated molecule at the $t$-th iteration. $F(x)$ is the set of constraints related to target properties. $\mathcal{N}(\hat{x}^{(t)})$ is the neighborhood set of molecules $\hat{x}^{(t)}$ \citep{fu2021differentiable}. Instead, Prabhumoye et al.~\cite{prabhumoye2018style} designed an style classifier to guide the style-specific generator to modify the style of the given text. 

As presented in Fig. \ref{fig:within_domain} (b), transfer learning-based transformation first trains the model in a large corpus of the unlabeled dataset, and then fine-tunes the model using the small-sized dataset $x_0'$ with labeled properties $c^*$ \citep{sudhakar2019transforming}. Adversarial training-based transformation, as the name suggests and shown in Fig. \ref{fig:within_domain} (c), borrows the adversarial loss to enforce the model to learn the distribution of the data with desired properties \citep{sood2018application, jin2018learning, fu2020core}. 
% \vspace{-4mm}
\begin{figure}[h]
\begin{center}
\includegraphics[width=0.7\textwidth]{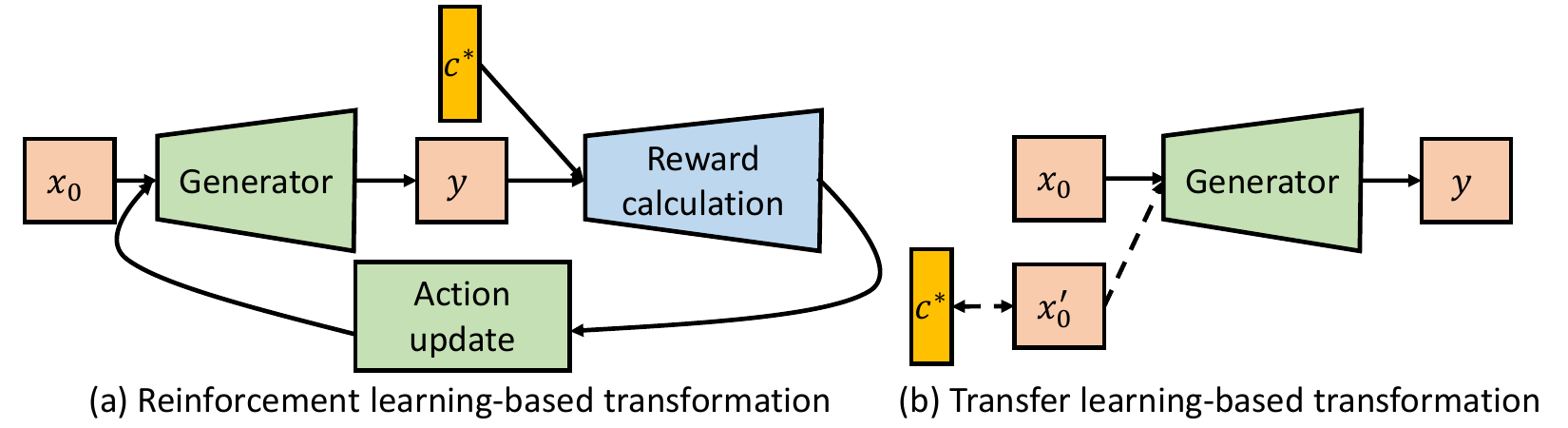}
% \vspace{-2mm}
\caption{Cross-modality transformation for controllable transformation from source data.}
\label{fig:cross_domain}
% \vspace{-5mm}
\end{center}
\end{figure}

\textbf{Cross-modality transformation} translates the source data $x_0$ to the target data $y$ while controlling its properties, where $x_0$ and $y$ are across different modalities. As shown in Fig. \ref{fig:transformation} (b) and Fig.~\ref{fig:cross_domain}, cross-modality transformation learns the distribution of the data with desired properties and generates from the source data directly without entailing information of properties during the generation process. The techniques used in this area roughly are aligned with those in peri-generation control discussed previously, except that it needs the source data as the input. In general, as shown in Fig. \ref{fig:cross_domain}, two techniques are borrowed to learn the distribution of data with desired properties to achieve cross-domain transformation: (1) reinforcement learning and (2) transfer learning. Reinforcement learning iteratively optimizes the reward calculated based on the target properties $c$ and other properties identified from the generated data $y$:
\begin{align}
y = & \argmax_{y} R(y, c, x_0) \quad s.t.\quad c \in c^*,
\label{eq:rwd_trans}
\end{align}
In this process, as shown in Fig.~\ref{fig:cross_domain} (a), the identifier that evaluates the generated data and calculates the rewards can be pre-trained \citep{liu2021reinforcement}. As illustrated by Fig. \ref{fig:cross_domain} (b), Transfer learning-based cross-modality transformation first trains the model in a large general dataset $x_0$. Then the model is fine-tuned by a small-sized dataset $x_{0}'$ with desired properties $c$ \citep{tits2019exploring}. As an example, Tits et al.~\citep{tits2019exploring} leveraged fine-tuning on a pre-trained deep learning-based Text-to-Speech (TTS) model to synthesize speech with specific emotions using a small-sized emotional dataset. Song et al.~\citep{song2018talking} generated the talking face video given an arbitrary speech clip and an arbitrary face image by encoding both image and speech via a recurrent adversarial network. In this work, the recurrent network first will be pre-trained for several epochs and then fine-tuned by adding discriminators to distinguish audio, image and lip reading.

\subsubsection{Steerable Transformation}\hfill\\
According to the source of the control signal, as shown in Fig.~\ref{fig:transformation} (2), steerable transformation techniques can be classified into two scenarios: (a) \textit{control via reference data} usually requires reference data in addition to the source data as the input to provide the control guidance; (b) \textit{control via latent space} controls the target properties via manipulating the latent variables corresponding to the properties of interest. 

The first scenario suffers from the challenge of obtaining the control signal as the reference data may come from different data modalities. Nevertheless, the second scenario, control via latent space, usually requires learning the relationship between the latent space and the properties of generated data, which can be challenging too.
% \vspace{-4mm}
\begin{figure}[h]
\begin{center}
\includegraphics[width=0.7\textwidth]{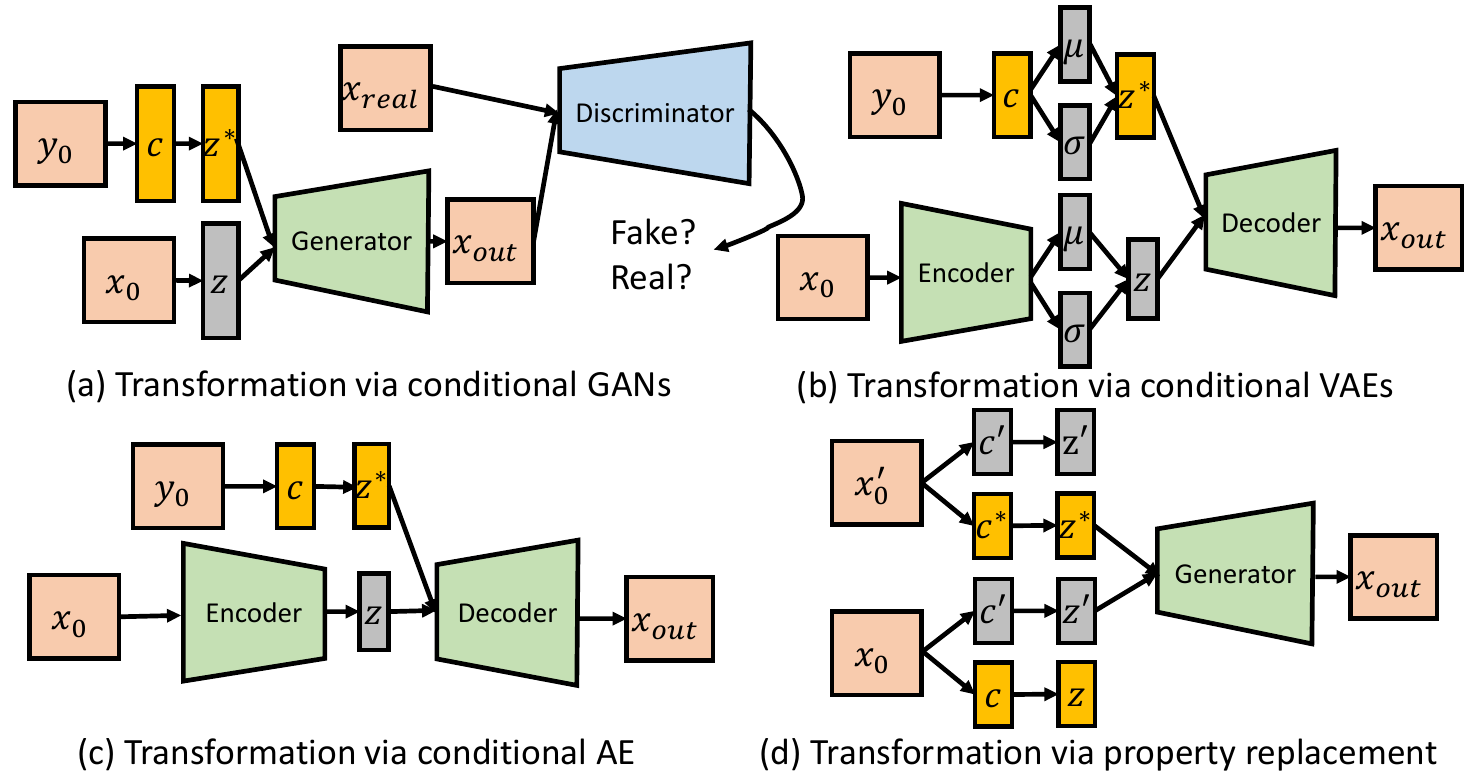}
% \vspace{-2mm}
\caption{Control via reference data for controllable transformation from source data.}
\label{fig:control_via_data}
% \vspace{-2mm}
\end{center}
\end{figure}

\textbf{Control via reference data}, as shown in Fig. \ref{fig:control_via_data}, requires both reference data and source data with the aim of altering the source data into the target data with some specific properties the same as those of reference data. For instance, the style or emotion of synthesized speech via text-to-speech (TTS) techniques can be controlled by adapting the style of a reference speech as the input to the generator \citep{kim2021expressive, cai2021emotion, hsu2018hierarchical, li2021controllable, liu2021reinforcement}. One can also transform the pose of a portrait into another pose provided by a reference portrait \citep{yang2021cpcgan}. Techniques for control via reference data usually have an encoder-decoder structure, in which target properties are extracted and encoded from the reference data while the data is generated conditional on the property representation along with the embedding of source data. These techniques, however, differ in how the target properties are fed into the generator. As shown in Fig. \ref{fig:control_via_data} (a), (b) and (c), some works extract and encode target properties $c$ from the reference data $y_0$ and feed its representation $z^*$ along with the embedding of source data into the generator. In this way the data is generated by conditioning on $c$. This strategy involves three basic types of techniques: (1) adversarial learning (i.e., Fig. \ref{fig:control_via_data} (a)); (2) variational inference (i.e., Fig. \ref{fig:control_via_data} (b)) and (3) conditional AE-based transformation (i.e., Fig. \ref{fig:control_via_data} (c)). The objective of adversarial learning-based models (i.e., Fig.~\ref{fig:control_via_data} (a)) can be formularized as:
\begin{equation}
    \mathcal{L} = \mathbb{E}_{x\sim p(x)}[\log D(x)] + \mathbb{E}_{z\sim p(z)}[\log (1-D(G(z, z^*\vert y_{0})))],
    \label{eq:condgan_multi}
\end{equation}
where $z$ is the representation of the source data, $z^*$ is the representation of target properties from the reference data $y_0$. Under the framework of adversarial learning, Chen et al.~\cite{chen2019hierarchical} generates talking face video first by learning audio and landmarks representations. CIAGAN anonymizes the face of the image based on the identity that is encoded from reference data via CNN \citep{maximov2020ciagan}. In this case, the objective function is aligned with Eq. \ref{eq:condgan_multi}. In addition to models based on adversarial learning, The framework based on the variational inference employed in control via reference data (i.e., Fig. \ref{fig:control_via_data} (b)) can be formularized as:
\begin{align}
    \mathcal{L} = -\mathbb{E}_{z\sim p(z\vert x), z^*\sim p(z^*\vert y_0)}[\log p_\theta(x\vert z, z^*)]+KL[q_{\phi}(z,z^*\vert x_0, y_0)\vert\vert p(z, z^*\vert y_0)],
    \label{eq:condvae_multi}
\end{align}
where the notation follows in the same role as in Eq. \ref{eq:condgan_multi}. Specifically, Chen et al.~\cite{chen2019controllable} controls the syntax of generated text by conditioning on the representation of a sentential exemplar under the framework of conditional VAE. Hsu et al.~\citep{hsu2018hierarchical} learns continuous attribute space of categorical observed labels, conditional on which the speech is generated from text. The learning objective of both models follows Eq. \ref{eq:condvae_multi}.

As illustrated in Fig. \ref{fig:control_via_data} (c), the third type of control via reference data-based technique is conditional AE, which main serves the controllable text-to-speech (TTS) transformation \citep{valle2020mellotron, kurihara2021prosodic, inoue2021model, kim2021expressive, cai2021emotion, li2021controllable, zhang2021ufc}. In this case the encoder and decoder can be an autoregressive model that handles sequential data such as text, audio and speech. In general, the framework of conditional AE aims to minimize the objective as follows: $\mathcal{L} = \vert x_0 - G(z, z^*\vert y_0) \vert$, where $G(z, z^*\vert y_0)$ corresponds to the decoder of the AE given the input of representations $z$ and $z^*$. The $L_1$ norm implemented above can be replaced with other forms of reconstruction loss such as MSE depending on different tasks. Under the framework of conditional AE, Valle et al.~\cite{valle2020mellotron} generates speech in a variety of styles conditioning on rhythm and continuous pitch contours from an audio signal or music score. Kurihara et al.~\cite{kurihara2021prosodic} proposed a model that controls prosodic features using phonetic and prosodic symbols as input for TTS transformation. Inoue et al.~\cite{inoue2021model} pre-trained emotional expression models using speech uttered by a particular person, and applied to another person to generate emotional speech with the person’s voice quality. StyleTagging-TTS extracts style embedding from reference speech via a reference encoder \cite{kim2021expressive} while Cai et al.~\cite{cai2021emotion} averages the style tokens’ weights for all reference audios to synthesize speech for this kind of audio. Li et al~\cite{li2021controllable} instead designed emotion classifier connected to the reference encoder to learn discriminative emotion embedding. UFC-BERT consists of modules of textual control, visual control, and preservation control to handle class labels and natural language descriptions, style transfer and preserving given image blocks, respectively \citep{zhang2021ufc}. 

By contrast, as shown in Fig. \ref{fig:control_via_data} (d), other works approach controllable transformation from source data in an interpretable manner by decomposing both source data and reference data into properties of interest, $c$ and $c^*$, and other non-relevant properties $c'$. Then the representation of $c$, $z$, is replaced with $z^*$, the representation of $c^*$. The generator takes $z^*$ and the representation $z'$ of non-relevant properties $c'$ as the input to generate target data \citep{liang2019pcgan}. Therefore, the objective of property replacement-based techniques can be concluded as follows:
\begin{align}
    \mathcal{L} =& - \mathbb{E}_{z, z_{x_0}'\sim p(z, z_{x_0}'\vert x_0)}[\log p_\theta(x_0\vert z, z_{x_0}')] - \mathbb{E}_{z^*, z_{x_0'}'\sim p(z^*, z_{x_0'}'\vert x_0')}[\log p_\theta(x_0'\vert z^*, z_{x_0'}')]\\\nonumber
    & - \mathbb{E}_{z_{x_0}'\sim p(z_{x_0}'\vert x_0), z^*\sim p(z^*\vert x_0')}[\log p_\theta(x\vert z_{x_0}', z^*)],
    \label{eq:prop_rep}
\end{align}
where $x_0$ is the source data and $x_0'$ is the reference data. $x_0$ and $x_0'$ are usually of the same type. $z_{x_0}'$ and $z_{x_0'}'$ correspond to the representation that is not relevant to the properties of interest of $x_0$ and $x_0'$, respectively. Minimizing the learning objective above learns latents of target properties and other properties of interest by the first two terms, then the data is generated based on the last term conditional on those latents. Yang et al.~\cite{yang2018unsupervised} transfers the style of the text by switching the style code of two texts and generating target text along with the content vector $z'$. Li et al.~\cite{li2018delete} transfers attributes of text by removing markers $z$ of the original attribute, then generating a new sentence conditioned on the remaining words and the target attribute $z^*$. 

\begin{figure}[h]
\begin{center}
\includegraphics[width=0.7\textwidth]{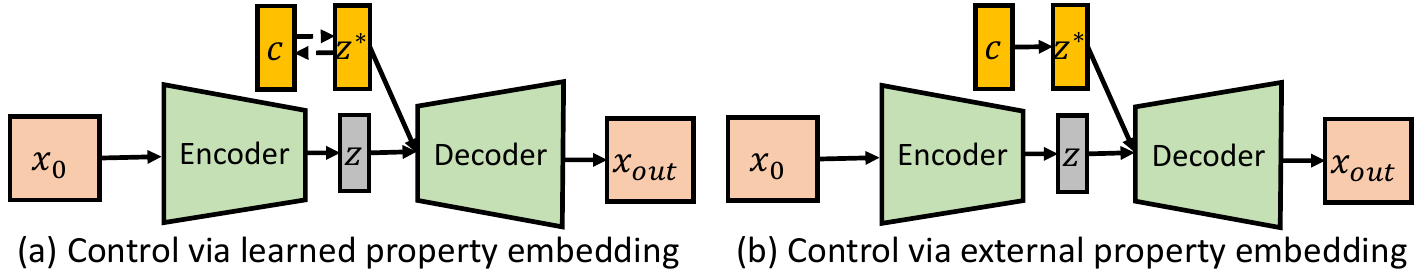}
% \vspace{-2mm}
\caption{Control via latent space for controllable transformation from source data.}
\label{fig:control_via_latent_space}
% \vspace{-4mm}
\end{center}
\end{figure}

\textbf{Control via latent space}. This type of method first maps the properties of interest either learned from data or provided externally into the latent space, and then controls the target properties of generated data via manipulating the latent variables corresponding to these target properties. Techniques of control via latent space usually have an encoder-decoder framework, where the information of (external) properties is embedded into the latent space. As shown in Fig. \ref{fig:control_via_latent_space}, according to how the properties serve as the input of the model, control via latent space-based techniques can be classified into two categories: (1) learned property embedding (Fig.~\ref{fig:control_via_latent_space} (a)) learns the properties from the source data via the encoder of the model \citep{ren2019fastspeech, engel2020ddsp, cui2021emovie, fabbro2020speech, qiao2020sentiment, shen2020interpreting, habib2019semi, thermos2021controllable, xia2021gan, zhu2021low} and (2) external property embedding (Fig. \ref{fig:control_via_latent_space} (b)) encodes the properties provided externally into the latent space from which new data is generated~\citep{guu2018generating, abdal2021styleflow, chang2021changing, john2018disentangled}. 

For learned property embedding-based techniques, the properties of interest are embedded into the latent space and usually disentangled from the representation of other information of the data: $\mathcal{L} = \vert x_0 - G(z, z^*\vert x_0) \vert$, where $z^*$ is the embedding of properties extracted from the source data $x_0$. Again, the $L_1$ norm can be replaced with other forms of distance measures if needed. For instance, FastSpeech controls the voice speed and prosody of generated speech by tuning the parameter of the length regulator as a part of the encoder \citep{ren2019fastspeech}. Differentiable Digital Signal Processing library integrates audio with classic signal processing elements, in which the fundamental frequency is learned from the source audio and embedded into the latent space \citep{engel2020ddsp}. EMOVIE achieves TTS transformation while controlling the emotion via the emotion embedding supervised by emotion labels \citep{cui2021emovie}. Fabbro et al.~\cite{fabbro2020speech} controls TTS systems by decomposing the spectrogram into control variables, such as amplitude envelope, harmonic distribution, and filter coefficients. SCTKG generates essays from given topics while controlling sentiment for each sentence by injecting the sentiment information into the generator \citep{qiao2020sentiment}. InterFaceGAN edits the face of the source image by interpolating disentangled latent semantics learned by GANs \citep{shen2020interpreting}. Habib et al.~\cite{habib2019semi} controls the prosody of the generated speech by manipulating the embedding of prosody learned by the TTS model. Disentangled anatomy arithmetic disentangles images into spatial anatomy (tensor) factors and accompanying imaging (vector) representations, and controls synthesized images by linearly combining the representation of corresponding properties \cite{thermos2021controllable}. GANs inversion inverts a given image back into the latent space of a pretrained GANs, then the image can be reconstructed from the latent space and the target properties can be controlled by intervening the style code on the latent space \citep{xia2021gan}. Low-rank subspaces enable more precise control of image generation via GANs by relating the latent space to the image region with the Jacobian matrix and then using low-rank factorization to discover steerable latent subspaces \citep{zhu2021low}. 

In addition to extracting the properties from the source data during the training process, external property embedding employs the external control signal, such as different property codes, and embeds the property information into the latent space \citep{guu2018generating, abdal2021styleflow, chang2021changing, john2018disentangled}: $\mathcal{L} = \vert x_0 - G(z, z^*\vert c) \vert$, where $z^*$ is the embedding of the external properties $c$. To distinguish the external property embedding-based techniques from the control via reference data-based techniques (Fig. \ref{fig:control_via_data}), control via reference data-based techniques first extract these properties from the reference data whereas external property embedding-based methods encode only the properties of interest. Specifically, the prototype-then-edit model generates a sentence by sampling a random example from the training set and then editing it using an edit vector encoding the type of edit to be performed \citep{guu2018generating}. StyleFlow borrows conditional continuous normalizing flows in the GANs latent space conditioning on attribute features \citep{abdal2021styleflow}. Attribute-controlled editing on images is achieved by the conditional exploration of entangled latent spaces \citep{abdal2021styleflow}. Chang et al.~\cite{chang2021changing} can produce a set of candidate topics by predicting the centers of word clusters, of which a user can select a subset to guide the generation of the text. John et al.~\cite{john2018disentangled} explicitly builds a classifier on the style space that predicts the style label, and combines the style representation with the source text representation to generate text with the desired styles. 

\section{Applications and Benchmark Datasets}
\label{sec:appl}
Controllable deep data generation has been gaining increasing attention of the community from various domains, enabling a broad range of applications including applications in molecule synthesis and optimization, protein design, image editing, text style transfer, and emotional speech generation. In addition, more and more datasets are becoming available to the community to train robust and expressive models for controllable deep data generation to advance the development of this domain. This section provides details of different techniques used for each application domain as well as the representative datasets used in this domain. More detailed introduction of benchmark datasets can be found in Appendix Section C.

\subsection{Molecular Synthesis and Optimization}
Molecules can be represented by several different data modalities, from sequence of characters~\citep{anderson1987smiles}, graphs~\citep{de2018molgan} to point clouds~\citep{gebauer2019symmetry}, etc. Designing molecules with desired properties, such as molecular weight, ClogP values, drug-likeness (QED), etc., naturally aligns with the goal of controllable deep data generation. Molecular design has been existing as a tough problem because of the complex molecular structure, vast chemical space, and its relationship with biological properties. This field can be divided into two types of problems: (1) molecular synthesis, which corresponds to the \emph{de novo} molecular design that requires designed molecule to preserve desired properties; (2) and molecular optimization, which aims to optimize the structure of an existing molecule towards the direction that specific properties are improved. The conventional methods to approach the tasks usually rely on the expert-derived heuristics \citep{walters2020applications} and suffer from the daunting computational complexity \citep{elton2019deep}. The rapid advances in the development of deep learning models have been powering molecular design to generate molecules with desired biological and chemical profiles. 

Molecular synthesis refers to the \emph{de novo} generation of novel chemical structures that satisfy a set of property constraints \citep{meyers2021novo,griffiths2022gauche}. Deep learning-based approaches have been used by researchers to develop novel and expressive representation of molecular structures, coupled with the ability to reveal relationships with properties~\citep{xiong2019pushing, shen2021out}. Then the molecular structure can be generated from the representation using either one-shot~\citep{simonovsky2018graphvae} or autoregressive~\citep{you2018graphrnn} strategies. This process often coordinates with techniques for goal-oriented control, such as reinforcement learning~\citep{you2018graph, neil2018exploring}, or multi-objective optimization to control the properties of generated molecules~\citep{jin2020multi, lambrinidis2021multi}. Datasets that have been used for molecular optimization include QM9~\citep{de2018molgan, ma2018constrained, liu2021graphebm, du2020interpretable, samanta2020nevae}, ZINC250K~\citep{you2018graph, jin2018junction, ma2018constrained, li2018learning, putin2018reinforced, liu2021graphebm, du2020interpretable, shi2020graphaf, yang2020practical, samanta2020nevae} and ChEMBL~\citep{li2018learning, segler2018generating, jin2020multi, neil2018exploring}.

Molecular optimization can be viewed as a transformation problem that maps from the source molecule to the target molecule that has high structural similarity with the original one but better properties~\citep{guo2022graph}. This task has been surmounted by extracting the signature of the target molecule and encoding its discrete chemical structure into the continuous latent space, which allows optimization in the molecular space. The control of properties of generated molecules requires single-modality fixed transformation with various strategies such as goal-oriented transformation, including reinforcement learning~\citep{ahn2020guiding, zhou2019optimization} and multi-objective optimization~\citep{winter2019efficient}, adversarial training-based transformation against target properties~\citep{jin2018learning}, etc~\citep{aumentado2018latent, fu2021mimosa}. Datasets that have been borrowed for molecular optimization are aligned highly with those for molecular synthesis, including QM9~\citep{luo2021graphdf, zang2020moflow, gomez2018automatic}, ZINC250K~\citep{jin2018junction, luo2021graphdf, zang2020moflow, gomez2018automatic, jin2018learning, fu2021differentiable, fu2020core, ahn2020guiding, zhou2019optimization, aumentado2018latent, fu2021mimosa}, ChEMBL~\citep{xie2020mars, wang2021multi, ahn2020guiding, zhou2019optimization} and MOSES~\citep{luo2021graphdf}.

\subsection{Protein Design}
Protein performs diverse physiological functions in a cell mediated by its three-dimensional structure composed of the sequence of amino acids \citep{orengo1999protein}. Protein design is a fundamental challenge in biology because of the costs both financially and timewise and has vast applications in domains such as drug discovery. The goal of protein design is to arrange the sequence of amino acids that fold to a specific protein structure. Unlike text or graph, protein design requires two-fold information, sequence design and structure design. While the fact that the protein sequence solely determines the protein structure and the big success of AlphaFold2 in predicting protein structures from protein sequences~\citep{jumper2021highly}, many works have been developed to design new sequences of protein and utilize AlphaFold2 to produce 3D structures. Sequences can be determined by the probability of 20 amino acids on each residue on the sequences~\citep{wang2018computational} or generated via property mapping-based frameworks such as VAEs and GANs~\citep{ding2022protein}. For some specific proteins, structures could be very diverse so that sequence-structure co-design is needed via property concatenation-based deep generative models conditioning on the full structure of the protein~\citep{ingraham2019generative}. Datasets regarding the protein structure that are available for protein design include Protein Data Bank~\citep{wang2018computational, ingraham2019generative}, UniProt~\citep{ding2022protein}, UniRef50~\citep{ding2022protein}, UniParc~\citep{ding2022protein} and Pfam~\citep{ingraham2019generative}.

\subsection{Image Editing}
By reason of the rapid growth of the internet and digital capture devices, a huge volume of images has become available for public use that make the training of deep models possible. Meanwhile, the advances of deep learning provide powerful tools such as CNN to process image data. The image editing task aims to generate a new image from a source image by editing the contents of the source image under certain guidance while keeping other properties unchanged~\citep{zhan2021multimodal}. The input and output of the model for image editing are the images represented by the pixel matrix with multiple color channels. The popular properties to be controlled under this scenario include but are not limited to facial expression, age and illumination of human portraits, artistic style, angle, shape, and color of the object, etc. The most popular framework to manage image editing is property concatenation or mapping-based techniques under the adversarial training of GANs. For instance, GAN-Control embeds the control signal in the latent space and achieves the controllability via modifying the latent variables~\citep{shoshan2021gan}. Song et al.~\cite{song2018talking} generates the talking face using the input audio as the control signal. CPCGAN transforms the pose of a portrait into another pose provided by a reference portrait \citep{yang2021cpcgan}. In addition to adversarial training-based approaches, frameworks based on variational inference~\citep{kingma2014semi, sohn2015learning, plumerault2020controlling, locatello2019disentangling} or flow~\citep{liu2019conditional, lu2020structured, kingma2018glow} are also borrowed for image editing. Datasets that have been employed for image editing cover a wide range of types of image, including digit image such as MNIST~\citep{mathieu2016disentangling, ardizzone2019guided, kingma2014semi, sohn2015learning, liu2019conditional} and The Street View House Numbers (SVHN)~\citep{radford2015unsupervised, kingma2014semi}, and human character, such as CelebA~\citep{upchurch2017deep, harkonen2020ganspace, lu2020structured, liu2019conditional, lu2020structured, kingma2018glow}, Flickr-Faces-HQ (FFHQ)~\citep{harkonen2020ganspace, shoshan2021gan}, Helen~\citep{upchurch2017deep}, Labeled Faces in the Wild (LFW)~\citep{sohn2015learning} and Sprites~\citep{mathieu2016disentangling}, shape images such as dSprites~\citep{plumerault2020controlling, locatello2019disentangling}, 3D shapes~\citep{locatello2019disentangling} and ShapeNet~\citep{yang2021cpcgan}, fashion image such as DeepFashion~\citep{hu2018deep}, animals such as Caltech-UCSD Birds 200~\citep{sohn2015learning} and general scenes such as ImageNet~\citep{radford2015unsupervised, jahanian2019steerability, ardizzone2019guided, kingma2018glow} and LSUN~\citep{radford2015unsupervised, kingma2018glow}.
\subsection{Text Style Transfer}
Deep learning has enabled the rapid development of designing novel natural language process (NLP) systems of controlling over various stylistic attributes of generated texts. Specifically, text style transfer intends to transfer a text with one stylistic attribute (e.g., politeness, emotion, humor, etc.) to a sentence with another stylistic attribute~\citep{jin2022deep}. Text style transfer can borrow the control signal via reference data-based techniques under the scenario of steerable transformation. For instance, Yang et al.~\cite{yang2018unsupervised} proposed an unsupervised model for text style transfer that uses the language model to evaluate generated sentences. The control signal comes from reference sentences with specific styles. In addition, control via latent space-based models are also used. \cite{john2018disentangled} employed an encoder-decoder structure to disentangle the latent variables of style and contents in language models. The style of generated texts can be controlled by feeding latent variables of target styles into the generator. CTRL instead is a conditional transformer language model trained to condition on control codes that govern style~\citep{keskar2019ctrl}. Text style transfer has been conducted on a few datasets including Yelp~\citep{xu2020variational, li2018delete, yang2018unsupervised, john2018disentangled, sudhakar2019transforming, guu2018generating}, Amazon~\citep{xu2020variational, li2018delete, john2018disentangled, sudhakar2019transforming, keskar2019ctrl}, Wikipedia~\citep{keskar2019ctrl}, RottenTomatoes~\citep{ficler2017controlling}, Project Gutenberg~\citep{keskar2019ctrl},  OpenWebText~\citep{keskar2019ctrl}, CAPTIONS~\citep{li2018delete, sudhakar2019transforming}, EMNLP2017 WMT News~\citep{prabhumoye2018style, keskar2019ctrl} and One Billion Word~\citep{guu2018generating}.

\subsection{Emotional Speech Generation}
Text to speech (TTS), also known as speech generation, aims to synthesize natural speech from text~\citep{taylor2009text}. Further, the naturalness of speech generation relies in its human-like, natural-sounding voice with desired emotional expression generated by the model~\citep{liu2021reinforcement}. Recent advances of emotional speech generation are overwhelmingly contributed by the development of deep learning techniques, especially those designed for controllable transformation from source data, which can learn expressive feature representations to characterize the emotional attributes of the speech~\citep{ning2019review}. Control via reference data-based models is studied broadly to approach this task. For example, Liu et al.~\cite{liu2021reinforcement} proposed i-ETTS which borrows the reference audio to extract the emotion embedding. Then the speech is generated under the framework of reinforcement learning to ensure that it preserves the desired emotions. GST-Tacotron2 also borrows the reference audio sequence to provide control signal and feed the embedding of both emotion and text to the WaveNet vocoder to predict the mel-spectrogram used to synthesize a speech waveform \citep{kwon2019effective}. Bian et al.~\cite{bian2019multi} designed a model that inputs multiple reference audios that independently disentangle and control specific styles. By contrast, control via latent space-based models are also employed in this application domain. Inoue et al.~\cite{inoue2021model} proposed a deep neural network-based model to control the emotion expressiveness of synthesized speech by directly feeding the emotional embedding to an emotion additive model. Datasets that have been utilized for emotional speech generation cover a broad ranges of languages including English such as LJSpeech~\citep{tits2019methodology, tits2019exploring}, LibriSpeech corpus~\citep{tits2019methodology},  LibriTTS~\citep{tits2019methodology}, CMU-ARCTIC~\citep{tits2019methodology},  The Interactive Emotional Dyadic Motion Capture (IEMOCAP)~\citep{tits2019methodology, cai2021emotion}, Catherine Byers~\citep{habib2019semi}, The RECOLA Database~\citep{cai2021emotion} and Emotional Speech Database~\citep{liu2021reinforcement, cui2021emovie}, Chinese such as Baidu Speech Translation Corpus (BSTC)~\citep{bian2019multi} and Chinese professional actress~\citep{li2021controllable}, Japanese such as Japanese emotional speech database~\citep{henter2018deep} and French such as SIWIS French Speech Synthesis Database~\citep{tits2019methodology}.

Together with the applications introduced above, more applications of controllable deep data generation along with representative works are summarized in Appendix Section D.

\section{Opportunities and Challenges}
\label{sec:chal}
All in all, controllable deep data generation is a fast-growing domain as it has a wide range of applications across various domains with numerous models constantly developed. In this section, we highlight a few promising future opportunities in this domain, followed by potential challenges that remain to be studied further.

\subsection{Opportunities}
We highlight some promising future opportunities on controllable deep data generation as follows.
\begin{itemize}[leftmargin=*]
\item \textbf{Control multiple properties simultaneously}. As far as we know, most of the existing works designed for controllable deep data generation only control one property at a time. Though technically challenging, simultaneously controlling multiple properties are commonly desired in real-world applications. For instance, chemists may want to generate QACs that have a high MIC value while preserving a small molecular weight. The task of simultaneously controlling multiple properties can be formalized naturally under a multi-objective optimization framework to enforce the generated data to satisfy multiple property constraints. A few works have been done in this direction. Jin et al.~\citep{jin2020multi} perform multi-objective optimization on the latent space using the gradient from the property predictors in their testing process. Wang et al.~\citep{wang2021multi} approach the multi-objective optimization task for controllable molecule generation by combining conditional transformer and reinforcement learning through knowledge distillation. Some more challenging tasks, however, such as constraining the property within a range during multi-objective optimization, still are under-explored.

\item \textbf{Improve model interpretability}. The interpretability of the model for controllable deep data generation is critical for the community to understand data. For instance, chemists may want not only to generate QACs with the desired value of MIC but also to know which substructure of generated QACs contributes to the corresponding MIC value. Jin et al.~\citep{jin2020multi} achieved the interpretability by learning the property rationals (i.e., substructure of the molecule) first and generating molecules with the specific property via assembling its rationals. In addition to interpreting properties of generated data, how to synthesize these data also attracts the attention of the community. For example, chemists may not be satisfied simply by generating molecules that are supposed to preserve desired properties but also want to know how to synthesize these molecules in the lab. Although some studies have been done regarding chemical reaction prediction~\citep{fooshee2018deep, schwaller2021prediction, cova2019deep}, few works are combined with the controllable deep data generation to infer the concrete synthesis process of chemicals.
\item \textbf{Develop semi-supervised or unsupervised learning-based techniques}. Existing datasets available for controllable deep data generation usually are trained by supervised learning. These properties, however, may not satisfy all needs of the community to generate data with properties of interest. For example, although one might generate QACs with specific MIC values, they cannot control other properties not labelled in the dataset. The property annotations, moreover, usually require significant time and labour costs, which are expensive for the community to obtain. As a result, semi-supervised learning-based techniques serve as a powerful tool to utilize sparsely annotated data. Unsupervised learning-based methods can also be designed to capture automatically properties of interest from unlabeled data via some well-designed regularization~\citep{wang2020learning, mollenhoff2019flat, pati2021disentanglement}.

\item \textbf{Entangle with hard constraint on data validity}. Data in some domains require strong validity constraint on their structure or properties. For instance, in molecular graph, the number of chemical bonds around an atom cannot exceed the numerical valency of the atom. Also, the combination of words in an sentence should follow specific grammar. Techniques for controllable deep data generation may learn from traditional methods to involve constraints of the domain knowledge when generating data with desired properties. Furthermore, these domain-specific constraints may be helpful in guiding the generation process.

\item \textbf{Generate out-of-sample data}. Current deep learning techniques for controllable deep data generation are highly data-driven, basically learn the distribution of the observed data and tend to generate similar data. The novelty of generated data, however, is highly desired but limited to the distribution of the observed data and hence resistant to exploring unseen subspace that may also have the potential of the desired value of properties. As a result, the extrapolation ability to generate out-of-sample data is critical when designing the model for controllable deep data generation. This task has been approached by some existing works via interpolating the latent space that controls target properties towards a pre-defined direction~\citep{radford2015unsupervised, plumerault2020controlling, jahanian2019steerability}. These methods, however, need to identify the pathway on the latent space where the properties are controlled and thus require extra effort and is difficult to obtain if the pathway is non-linear.
\end{itemize}

\subsection{Open challenges}
Despite numerous methods proposed for efficient and effective generation design, several open challenges still remain on controllable deep data generation.
\begin{itemize}[leftmargin=*]
\item \textbf{Lack of theoretical ground}. Existing frameworks for controllable deep data generation, e.g., conditional GANs and conditional VAEs, learn and sample from the distribution of the observational data while neglecting domain-specific theories that potentially can improve the model performance. By contrast, those domain-specific theories have been borrowed widely by traditional models to guide the optimization process of the model~\citep{lippow2007progress}. For instance, conventional models for protein design borrow the energy functions such as those for nucleic acids and their interactions with protein to examine large combinatorial collections of protein structure candidates~\citep{lippow2007progress}. In this way, generated proteins theoretically can be guaranteed to have low energy. As a result, how to contain and model those domain-specific theories into deep learning-based frameworks to improve the performance of controllable deep data generation remains to be explored further.

\item \textbf{Lack of interpretability}. Many works have performed well in optimizing the well-designed objective, while ignoring the interpretation of the design process. Specifically, some tasks conduct optimization on molecular structure only by searching structures that may cause specific molecular properties while putting less effort on the understanding of active groups that correspond to these properties. In addition, although generated molecules might own valid structures, chemists might be more interested in how to synthesize them in the lab. The synthesis pathways of these optimized molecules, however, still remain unknown or extremely hard to synthesize and lead to unrealistic applications.

\item \textbf{Difficulty in capturing and optimizing correlated properties}. Capturing and considering the correlation among properties during controllable deep data generation can deepen our understanding of data. To date, most of works only consider controlling independent properties of generated data, even though correlated properties are ubiquitous for data in the real world. For instance, given two correlated properties such as molecular weights and solubility, chemists may still want to increase the solubility of the generated molecules while keeping the molecular weights unchanged. Optimizing correlated properties is challenging, since this process may lead to the conflicting property constraints of the multi-objective optimization framework and thus call for developing more advanced optimization techniques from the community.

\item \textbf{Paucity of labelled dataset}. The property annotations of the datasets used to train the model for controllable deep data generation usually require enormous financial, time and human costs. For instance, the property of MIC of QACs that have been optimized to generate novel QACs needs to be measured manually in the lab~\citep{buffet2011effect}. The expensive annotation, moreover, may lead to other problems such as the small sample size of the available dataset (e.g., only 462 QACs are processed and contained in the QAC dataset), which is hard for the supervised training of techniques for controllable deep data generation such as those for goal-orientation transformation. As a result, a semi-supervised or unsupervised approach for controllable deep data generation is needed to learn and control the properties of small data.

\item \textbf{Difficulty in automatic validation of properties of generated data}. Automatic validation of properties of generated data is critical yet challenging for some domains. For instance, the task of music generation often relies on manual validation, which requires a human evaluator to score and rank the generated data based on some pre-determined rules. Validation of the bio-activities of generated molecules instead requires conducting lab experiments. These downstream evaluations hence suffer from the high cost of time and money, as well as the noise introduced by the subjective evaluation or the batch effect of lab experiments.

\end{itemize}

\section{Conclusion}
\label{sec:con}
This survey has provided a comprehensive review of existing techniques developed for controllable deep data generation. The first consideration is potential challenges and preliminaries. Then, the problem of controllable deep data generation receives a formal definition, followed by a novel taxonomy of techniques designed for controllable data generation. The evaluation metrics in this specific domain, moreover, are summarized in terms both of data quality and property controllability. After that, a systemic review and comparison of techniques involved in the taxonomy are conducted, followed by their exciting applications and benchmark datasets borrowed in the domain of controllable deep data generation. Finally, the survey concludes with the highlight of promising future directions and five potential challenges in this domain. We believe this survey will pave the way for the future study of controllable deep data generation in many disciplines.

\bibliographystyle{unsrt}
\bibliography{sample-base}

\newpage
\appendix
\section{Preliminaries}
\label{sec:prelim}

This section introduces the common building blocks for controllable deep data generation, including GANs, Auto-encoder (AE), VAEs, normalizing flows, diffusion models, etc., as well as their roles in data generation. Let $x$ be the data object and $z$ be the latent variable if involved. Those common building blocks are introduced as follows.

\subsection{Generative Adversarial Nets}
Generative Adversarial Nets (GANs) was originally proposed for drawing new samples from data distribution and enriching the dataset for more efficient training of deep learning models. Specifically, GANs learns to sample realistic objects via distinguishing real objects from synthetic objects~\cite{goodfellow2014generative}. Two main components are generator and discriminator, contained in a GANs-based model. The generator aims to generate from random noise realistic objects indistinguishable by the discriminator from real objects. The discriminator aims to distinguish those generated objects from the real ones. Overall, the generator and the discriminator play a min-max game. Let $G$ be the generator and $D$ be the discriminator, and both of them are parameterized by neural networks. The objective of GANs can be written as follows.
\begin{equation}
\min_G \max_D V(G,D) = \mathbb{E}_{x\sim p(x)}[\log D(x)] + \mathbb{E}_{z\sim p(z)}[\log(1-D(G(z))].
\end{equation}
\subsection{Auto-encoders and Variational Auto-encoders}
Auto-encoder (AE) consists of an encoder, a latent space mapped from the encoder and a decoder to recover the data from the latent space~\cite{liou2014autoencoder}. AE is trained by minimizing a common objective function that measures the distance between the reconstructed and original data. Let $dist()$ represent the measure of distance which can be $L_1$ norm, mean squared error, etc. Let $Dec$ be the decoder that reconstructs the data from the latent space obtained from the encoder. Then the objective function of AE can be formularized as below.
\begin{gather}
    \mathcal{L}(x, z) = dist(x, Dec(z)),
\end{gather}

Variational Auto-encoders is extended from AE instead by learning the distribution of data via encoder based on variational inference from which similar data can be sampled by the decoder~\cite{kingma2014auto}. Specifically, VAEs aims to learn a joint distribution between the latent space $z\sim p(z)$ and samples $x\sim p(x)$. Due to the intractability of the marginal likelihood of $p(x)$, the encoder approximates the posterior $p(z\vert x)$ via the inference network $q_{\phi}(z\vert x)$ to encode $x$ into latent space $z$. Then the decoder is defined as a generative distribution $p_{\theta}(x\vert z)$. Here $\phi$ and $\theta$ are trainable parameters of encoder and decoder, respectively. 

VAEs maximize a marginal likelihood of the data as $\max\limits_{\phi, \theta}\mathbb{E}_{q_{\phi}(z\vert x)}[\log p_{\theta}(x\vert z)]$. Let $D_{KL}$ be the operation of Kullback–Leibler (KL) divergence. The marginal likelihood of samples can be rewritten as:
\begin{gather}
    \log p_{\theta}(x\vert z) = \mathcal{L}(x, z; \phi, \theta) + D_{KL}(q_{\phi}(z\vert x)\Vert p(z)),\nonumber
\end{gather}
where the second term measures the distance between the true and the approximated posterior. For the first term, rather than directly performing the maximum likelihood estimation, VAEs optimize the tractable evidence lower bound (ELBO):
\begin{gather}
    \mathcal{L}(x, z; \phi, \theta) = \mathbb{E}_{q_{\phi}(z\vert x)}[p_{\theta}(x\vert z)] - D_{KL}(q_{\phi}(z\vert x)\Vert p(z)).\nonumber
\end{gather}
In practice, to make the above optimization process tractable, a reparameterization trick is usually adopted. We assume a simple prior distribution of $p(z)$ as $\mathcal{N}(\mathbf{0}, \mathbf{I})$. In this process, $z\sim q_{\phi}(z\vert x)$ is parameterized as Gaussian with a differentiable transformation of a noise variable $\epsilon \sim \mathcal{N}(\mathbf{0}, \mathbf{I})$ by $z=\mu+\sigma\odot\epsilon$, where $\mu$ and $\sigma$ are the output of the encoder.

\subsection{Normalizing Flows}
Normalizing flows models the exact likelihood of the data by building an invertible mapping between the data $x$ and the learned latent space $z$. Let $f$ be the function of the encoder $z=f(x)$ and accordingly $f$ is invertible such that $x=f^{-1}(z)$. The distribution of the latent variables can be determined by the change of variable formula as follows:
\begin{equation}
p(z) = p(x)|\det\frac{\partial f^{-1}}{\partial z}| = p(x)|\det \frac{\partial f^{-1}}{\partial z}|,
\end{equation}
where the density of the data distribution $p(x)$ can be obtained by a chain of $K$ transformations from the random variable $z_0$ (Eq.~\ref{eq:nfzk}) and the joint likelihood of data can be represented as below (Eq.~\ref{eq:nfpx}).
\begin{equation}
x = z_K = f_K \circ \cdots \circ f_2 \circ f_1(z_0),
\label{eq:nfzk}
\end{equation}
\begin{equation}
\log p(x) = \log p_0(z_0) - \sum_{k=1}^{K}\log|\det \frac{\partial f_k}{\partial z_{k-1}}|
\label{eq:nfpx}
\end{equation}
Designing invertible transformations is a nontrivial problem, however, and requires the input and latent dimension of the model to be equal. 

\subsection{Reinforcement Learning}
Reinforcement learning (RL) is a common technique employed for goal-oriented data generation, where target properties keep being improved via taking a series of actions to maximize designed reward function in a prescribed environment. The environment can be a Markov decision process for molecular modification \cite{zhou2019optimization}, or that carrying out actions that obey the given rules of chemistry in the molecule generation process \cite{you2018graph}. RL-based methods try to fit a function $Q(s, a)$ that predicts the future rewards of taking an action $a$ on state $s$. A decision is made by choosing the action $a$ that maximizes the Q function, which leads to larger future rewards. For a policy $\pi$, let $r_n$ be the reward at step $n$ and then we can define a value of an action $a$ on a state $s$ as:
\begin{equation}
Q^{\pi}(s, a) = Q^{\pi}(m, t, a)=\mathbb{E}_{\pi}[\sum_{n=t}^{T}r_n].
\end{equation}
The optimal policy can be then defined as $\pi^*(s)=arg\,max_a Q^{\pi^*}(s,a)$.

% \subsection{Auto-regressive Models}
\subsection{Generative Diffusion}
\subsubsection{Diffusion Models}\hfill\\
Diffusion models is the generative model that contains a forward process to add successively Gaussian noise to the data and a reverse process to recover the data by transforming the noise back into a sample from the target distribution~\cite{ho2020denoising}. Specifically, until the step $T$, the forward process can be formularized as follows based on the Markov assumption. Suppose $\beta_1$,...,$\beta_T$ are the variance that can be either fixed or learned:
\begin{equation}
q(x_{1:T}\vert x_0) = \prod_{t=1}^Tq(x_t\vert x_{t-1}) = \prod_{t=1}^T\mathcal{N}(x_t;\sqrt{1-\beta_t}x_{t-1}, \beta_t I).
\end{equation}
The above formula shows that the forward process can be set to the product of conditional Gaussians when the noise level is sufficiently low. Accordingly, the reverse process starts from the initial Gaussian noise $p(x_T)=\mathcal{N}(0, I)$ and learns the joint distribution $p(x_{0:T})$. Let $\theta$ be the model parameter independent of time:
\begin{equation}
p_{\theta}(x_{0:T}) = p(x_T)\prod_{t=1}^Tp_{\theta}(x_{t-1}\vert x_t)=p(x_T)\prod_{t=1}^T\mathcal{N}(x_{t-1}; \mu_{\theta}(x_t, t), \Sigma_{\theta}(x_t, t)).
\end{equation}
The model is trained by finding the reverse Markov transitions that maximize the likelihood of the training data. Define $\epsilon_{\theta}(x_t, t)$ as the noise component of the sample $x_t$ and $\epsilon$ as a random noise. Then the loss function can be further derived, simplified and formularized as:
\begin{equation}
\mathcal{L}=\mathbb{E}_{t, x_0, \epsilon}[\vert\vert\epsilon - \epsilon_{\theta}(x_t, t)\vert\vert_2^2].
\label{eq:ddpm_l}
\end{equation}
Once the diffusion model is trained, we can generate data by passing the random noise through the learned denoising process.
\subsubsection{Score-based Generative Diffusion}\hfill\\
Additionally, score-based generative diffusion instead models the score function $s_\theta(x)=\nabla_x \log p(x)$ to avoid computing intractable $\log p_\theta(x)$ during the diffusion process~\citep{song2020score}. Unlike the conventional diffusion model that adds discrete noise to the data, score-based diffusion constructs the diffusion process by a continuous time variable $t\in[0, T]$. The diffusion process can be formulated as the solution to an stochastic differential equation (SDE):
\begin{align}
    dx = f(x, t) dt + g(t)dw,
\end{align}
where $w$ is the standard Wiener process. $f(\cdot, t)$ is a vector-valued function and $g(\cdot)$ is a scalar function serving as the diffusion coefficient of $x_t$.

Then the reverse process starts from $x_T$ to obtain $x_0$, which is given by the reverse SDE:
\begin{align}
    dx=[f(x, t) - g(t)^2\nabla_x\log p(x_t)]dt + g(t)d\bar{w},
\end{align}
where $\bar{w}$ is a standard Wiener process also. If the score of each marginal distribution at $t$, $\nabla_x\log p(x_t)$, is know, we derive the reverse diffusion process above.

To estimate the score $\nabla_x \log p(x_t)$, we can train a time-dependent score based model $s_\theta(x, t)$ following~\cite{song2020score}:
\begin{align}
    \mathcal L=\mathbb E_{t, x_{0:T}}[\lambda(t)\vert\vert s_\theta(x_t, t)-\nabla_{x_t}\log p(x_t\vert x_0)\vert\vert_2^2].
\end{align}
Once the model is trained, we can use score-based Markov chain Monte Carlo (MCMC), such as Langevin MCMC, to sample data directly from the score function or the probability.

\section{Experimental comparison and analysis}
\label{sec:exp}
In this section, we showcase some experimental results of controllable deep data generation regarding molecule design, image synthesis, text style transfer, and controllable text-to-speech synthesis.

\subsection{Molecule Design}
% Molecule generation normally involves two goals: (1) generating valid and novel molecules, and (2) controlling the properties of generated molecules. In this review, our focus is on controllable deep molecule generation so we ignore the evaluation for the first task. 
Controllable molecule design aims to generate valid and novel molecules with desired properties and can be divided into two types of tasks: (1) controllable molecule synthesis and (2) molecule optimization. 

Controllable molecule synthesis refers to generating novel molecules with various structures but with optimal properties. Two commonly used molecular properties as optimization targets are logP and QED~\cite{bickerton2012quantifying}. Penalized logP is a combined objective of logP, synthesis accessibility and ring size. QED is a quantitative estimation of drug-likeness. Then we evaluate the ability of the model in generating molecules with optimal properties. We report the generated molecules with top-3 molecular properties here. Based on the results presented in Table~\ref{tab:mol_prop_opt}, GraphDF achieves the best optimal property values on both plogP, which outperforms the second-best model GraphAF by 1.83 on average. GCPN, MoFlow and GraphAF have the comparable results as GraphDF on QED. Generally speaking, the results indicate that the flow-based models (i.e., GraphAF, MoFlow and GraphDF) perform better than others on QED by outperforming JT-VAE and GCPN by 0.14 on average. Those flow-based models employ the invertible function to reconstruct molecules while entangling the latent variables in the middle with target properties. This mechanism could enhance stronger constraints between properties and reconstructed data than its counterparts as it allows exact likelihood calculation of data while others may need to approximate intractable components. Models that are based on reinforcement learning on property optimization (i.e., GCPN, GraphAF and GraphDF) are observed to have a better performance on plogP. This improvement is the result of reinforcement learning-based methods that can search on the large data space for the optimal properties, which can generate better out-of-sample data (i.e., extreme plogP value) than its counterparts.

% ~\cite{gao2022sample} 

\begin{table}[htb]
    \centering
    \caption{Molecule synthesis}
    \begin{adjustbox}{max width=0.5\linewidth}
    \begin{tabular}{cc|c|ccccc}
    \toprule
    \hline
         \multicolumn{2}{c|}{Property} & ZINC & JT-VAE & GCPN & GraphAF & MoFlow & GraphDF \\\hline
         \multirow{3}{*}{plogP}&1$_{st}$& 4.52 & 5.3 & 7.98 & 12.23 & 4.96 & \textbf{13.7}\\
         ~&2$_{nd}$ & 4.3 & 4.93 & 7.85 & 11.29 & 4.88 & 13.18\\
         ~&3$_{rd}$& 4.23 & 4.49 & 7.80 & 11.05 & 4.75 & 13.17 \\\cline{1-8}
         \multirow{3}{*}{QED}&1$_{st}$& 0.948 & 0.925 & \textbf{0.948} & \textbf{0.948} & \textbf{0.948} & \textbf{0.948}\\
         ~&2$_{nd}$ & 0.948 & 0.911 & 0.947 & \textbf{0.948} & \textbf{0.948} & \textbf{0.948} \\
         ~&3$_{rd}$& 0.948 & 0.910 & 0.946 & 0.947 & \textbf{0.948} & \textbf{0.948}\\
         \hline
    \end{tabular}
    \end{adjustbox}
    \label{tab:mol_prop_opt}
\end{table}

Molecule optimization refers to generating molecules with similar structures as the input reference molecule but with improved properties. To compare the above-mentioned methods on molecule optimization task, we report their performance on the improvement of penalized logP for 800 molecules with the lowest penalized logP property from ZINC250K dataset. Two distinct sets of 800 molecules are used in the literature, however. We name the dataset on which JT-VAE~\cite{jin2018junction}, GCPN~\cite{you2018graph} and GraphDF~\cite{luo2021graphdf} are tested as A, and the dataset on which GraphAF~\cite{shi2020graphaf}, MoFlow~\cite{zang2020moflow} and GraphDF~\cite{luo2021graphdf} are tested as B. Full results appear in Table \ref{tab:mol_cons_opt}. $\delta$ denotes the structural similarity of the generated molecules with the base molecules. We carried out additional experiments to evaluate their performance on the alternative dataset reported in the original paper, and results from our additional experiments are marked with an asterisk ($*$) at the end of each line. Overall, GraphDF achieves the largest improvement on both datasets and exceeds the second-best model GraphAF by 0.34 and 0.6 on average of the improvement on data A and B, respectively. GraphDF is particularly designed to handle the discrete structure of the graph (molecule) via normalizing flow-based strategy and naturally map the discrete latent variables to graph (molecule) structures. Noticeably, the performance of GraphDF and GraphAF are comparable regarding the success rate since GraphAF is also a flow-based approach to control properties via reinforcement learning and they both are normalizing, flow-based models, which can achieve stronger constraints among properties and reconstructed data by allowing exact calculation of likelihood based on data.
\begin{table}[]
\centering
\caption{Molecule optimization on plogP}
\label{tab:mol_cons_opt}
\begin{adjustbox}{max width=\linewidth}
\begin{tabular}{ll|llll|llll}
\hline
\multicolumn{2}{l|}{Dataset}                                & \multicolumn{4}{l|}{A} & \multicolumn{4}{l}{B} \\ \hline
\multicolumn{2}{l|}{Delta}                                  & 0.0 & 0.2 & 0.4 & 0.6 & 0.0  & 0.2 & 0.4 & 0.6 \\ \hline\hline
\multicolumn{1}{l|}{\multirow{3}{*}{JT-VAE}}  & Improvement & $1.91 \pm 2.04$    & $1.68 \pm 1.85$    & $0.84 \pm 1.45$    & $0.21 \pm 0.71$    & $2.89\pm 2.10$     & $1.83\pm 1.69$    & $1.07\pm 1.14$    & $0.45\pm 0.60$    \\
\multicolumn{1}{l|}{}                         & Similarity  & $0.28 \pm 0.15$    & $0.33 \pm 0.13$    & $0.51 \pm 0.10$    & $0.69 \pm 0.06$    & $0.13\pm 0.09$     & $0.29\pm 0.10$    & $0.54\pm 0.14$    & $0.75\pm 0.14$    \\
\multicolumn{1}{l|}{}                         & Success     & 97.5\%    & 97.1\%    & 83.6\%    & 46.4\%    & 68.9\% *     & 52.8\% *    & 30.7\% *    & 15.4\% *    \\ \hline
\multicolumn{1}{l|}{\multirow{3}{*}{GCPN}}    & Improvement & $4.20 \pm 1.28$    & $4.12 \pm 1.19$    & $2.49 \pm 1.30$    & $0.79 \pm 0.63$    & $-$     & $-$    & $-$    & $-$    \\
\multicolumn{1}{l|}{}                         & Similarity  & $0.32 \pm 0.12$    & $0.34 \pm 0.11$    & $0.48 \pm 0.08$    & $0.68 \pm 0.08$    & $-$     &  $-$   & $-$    & $-$    \\
\multicolumn{1}{l|}{}                         & Success     & \textbf{100\%}    & \textbf{100\%}    & \textbf{100\%}    & \textbf{100\%}    & $-$     &  $-$   & $-$    & $-$    \\ \hline
\multicolumn{1}{l|}{\multirow{3}{*}{GraphAF}} & Improvement & $5.22\pm 1.47$    & $5.03\pm 1.35$    & $3.74\pm 1.21$    & $2.05\pm1.02$    & $13.13\pm 6.89$     & $11.90 \pm 6.86$    & $8.21\pm 6.51$    & $4.98\pm 6.49$    \\
\multicolumn{1}{l|}{}                         & Similarity  & $0.32\pm 0.12$    & $0.34\pm 0.11$    & $0.47\pm0.08$    & $0.66\pm0.05$    &  $0.29 \pm 0.15$    & $0.33 \pm 0.12$    & $0.49\pm0.09$    & $0.66\pm0.05$    \\
\multicolumn{1}{l|}{}                         & Success     & \textbf{100\%} *    & \textbf{100\%} *    & \textbf{100\%} *    & 98.4\% *    & \textbf{100\%}     & \textbf{100\%}    & 99.88\%    & 96.88\%    \\ \hline
\multicolumn{1}{l|}{\multirow{3}{*}{MolFlow}} & Improvement & $3.35\pm 1.39$    & $3.25\pm 1.35$    & $3.14 \pm 1.22 $    & $2.84 \pm 1.03$    & $8.61\pm 5.44$     & $7.06 \pm 5.04$    & $4.71\pm4.55$    & $2.10\pm2.86$   \\
\multicolumn{1}{l|}{}                         & Similarity  & $0.57\pm 0.24$    & $0.61\pm 0.20$    & $0.67\pm 0.15$   & \textbf{0.75 $\pm$ 0.11}    & $0.30 \pm 0.20$     & $0.43 \pm 0.20$    & $0.61\pm0.18$    & \textbf{0.79$\pm$0.14}    \\
\multicolumn{1}{l|}{}                         & Success     & 98.9\% *    & 98.1\% *    & 94.4\% *   & 79.6\% *    & 98.88\%     & 96.75\%    & 85.75\%    & 58.25\%    \\ \hline
\multicolumn{1}{l|}{\multirow{3}{*}{GraphDF}} & Improvement & \textbf{5.93$\pm$ 1.97}    & $5.62 \pm 1.65$    & $4.13\pm 1.41$    & $1.72\pm 1.15$    & \textbf{14.15$\pm$ 6.86}     & $12.77\pm 6.59$    & $9.19\pm 6.43$    & $4.51\pm 5.80$    \\
\multicolumn{1}{l|}{}                         & Similarity  & $0.30 \pm 0.12$    & $0.32\pm 0.10$    & $0.47\pm0.07$    & $0.67\pm 0.06$    & $0.29 \pm 0.13$     & $0.32\pm 0.11$    & $0.48\pm 0.08$    & $0.65\pm 0.05$    \\
\multicolumn{1}{l|}{}                         & Success     & \textbf{100\%}    &  \textbf{100\%}   & \textbf{100\%}    & 93\%    & \textbf{100\%}     & \textbf{100\%}    & 99.63\%    & 92.13\%    \\ \hline
\end{tabular}
\end{adjustbox}
\end{table}

\subsection{Image synthesis}
Controllable deep image synthesis generates images that preserve desired properties. Borrowing the FFHQ dataset, we evaluate the human face editing task for image synthesis by collecting and summarizing results from Abdal et al.~\cite{abdal2021styleflow} in Table~\ref{tab:faceedit}. The human face editing task requires the model to edit one attribute at a time while keeping other attributes unchanged. The results in Table~\ref{tab:faceedit} show how sequential editing of three attributes (i.e., light, pose, and expression) performed on the image. Based on the results, StyleFlow outperforms the other models, including Image2StyleGAN, InterfaceGAN and GANSpace, in preserving the identity of the image by 0.186, 0.052 and 0.667 on average regarding Euclidean distance, cosine similarity, and accuracy, respectively. This superior performance also validates the power of the flow-based model on controllable deep data generation on image editing-related tasks. It is worth noting that although flow-based models have superior performance regarding the property control, nonetheless they can suffer from the slow sampling speed and poor scaling properties~\cite{bond2021deep}. One exception is GANSpace that beats others when editing the expression of the image~\cite{harkonen2020ganspace}. 

\begin{table}[htb]
    \centering
    \caption{Identity preservation achieved by different methods as evaluated by a SOTA face classifier; Es, Cs and Acc represent Euclidean distance, cosine similarity and accuracy, respectively.}
    % \label{tab:mol_cons_opt}
    \begin{adjustbox}{max width=0.7\linewidth}
    \begin{tabular}{c|cc|cc|cc|ccc}
    \toprule
         \multirow{2}{*}{Model} &  \multicolumn{2}{c|}{Light} & \multicolumn{2}{c|}{Pose} & \multicolumn{2}{c|}{Expression} & \multicolumn{3}{c}{All} \\
         ~& Es & Cs & Es & Cs & Es & Cs & Es & Cs & Acc\\
         \hline
         Image2StyleGAN & 0.633 & 0.910 & 0.748 & 0.877 & 0.534 & 0.941&0.774 &0.870 &0.000\\
         \hline
         InterfaceGAN &  0.508 & 0.945 & 0.532 & 0.940 &0.509 &0.946 & 0.690&0.895 &0.300\\
         \hline
         GANSpace & 0.524 & 0.942 & 0.526 & 0.939 &\textbf{0.359} &\textbf{0.973} &0.681 &0.902 &0.550\\
         \hline
         StyleFlow & \textbf{0.394} & \textbf{0.963} & \textbf{0.400} & \textbf{0.966} & 0.388 &0.967 &\textbf{0.529} &\textbf{0.941} &\textbf{0.950}\\
         \hline
    \end{tabular}
    \end{adjustbox}
    \label{tab:faceedit}
\end{table}

\subsection{Text style transfer}
We adapted results regarding the quantitative evaluation of text style transfer from Sudhakar et al.~\cite{sudhakar2019transforming}. Three datasets, Yelp, Amazon and Captions, assist in evaluating six models including StyleEmbedding~\cite{fu2018style}, MultiDecoder~\cite{fu2018style}, CrossAligned~\cite{shen2017style}, DeleteAndRetrieve~\cite{li2018delete}, Blind Generative Style Transformer~\cite{sudhakar2019transforming} and Guided Generative Style Transformer~\cite{sudhakar2019transforming}. As shown in Table~\ref{tab:text_trans}, DeleteAndRetrieve beats other models on Yelp and Captions dataset by 35.6$\%$ on average regarding the target style accuracy as it achieves the style transfer in a delete-and-retrieve manner by deleting phrases associated with the original attribute and retrieving new phrases associated with the target attribute. CrossAlighned performs better on Amazon dataset, which adopts a cross-alignment strategy by sharing the latent semantic space between two sentences with different styles. Specifically, CrossAlighned aligns generated sentences with the sentence of the target style at the distributional level to let the generated sentence preserve the target style. Here the target style accuracy is obtained from the pre-trained model using FastText~\cite{joulin2016bag}.

\begin{table}[htb]
    \centering
    \caption{Results on text style transfer}
    \begin{adjustbox}{max width=0.85\linewidth}
    \begin{tabular}{c|cc|cc|cc}
    \toprule
    \hline
         \multirow{2}{*}{Model} &  \multicolumn{2}{c|}{Yelp} & \multicolumn{2}{c|}{Amazon} & \multicolumn{2}{c}{Captions} \\
         ~& Perplexity & Accuracy & Perplexity & Accuracy & Perplexity & Accuracy \\\hline
         StyleEmbedding &115.9 &8.6\% & 129.8& 45.5\%& 80.3& 51.0\%\\
         \hline
         MultiDecoder &205.6 & 46.8\% &122.5 &71.8\% & 40.5 & 51.3\%\\
         \hline
         CrossAligned & 72.8 & 72.7\% & \textbf{30.1} &\textbf{83.1\%} &\textbf{10.1} & 50.8\%\\
         \hline
         DeleteAndRetrieve &90.0 &\textbf{89.3\%} &42.2 &50.9\% &28.8 &\textbf{67.5\%}\\
         \hline
         Blind Generative Style Transformer &\textbf{38.6} &87.3\% &55.2 & 60.0\% &28.9 &56.0\% \\
         \hline
         Guided Generative Style Transformer &64.4 &78.3\% &171.0 &57.6\% & 45.0&52.3\% \\
         \hline
    \end{tabular}
    \end{adjustbox}
    \label{tab:text_trans}
\end{table}

\subsection{Controllable text-to-speech synthesis}
We summarized the evaluation results of controllable deep text-to-speech synthesis from existing published works. The mean opinion score (MOS)-based evaluation on audio quality was performed by Ren et al.~\cite{ren2019fastspeech} on the LJSpeech dataset by at least 20 native English listeners, and the results are summarized in Table \ref{tab:tts}. Based on the results, Transformer TTS~\cite{ping2018deep} has the best MOS score among all other models involved in the evaluation by 0.51 on average regarding MOS. The transformer-based models are more suitable to handle the sequential nature of the generated speech data. The performances of Tacotron 2~\cite{8461368}, Transformer TTS and FastSpeech~\cite{ren2019fastspeech} are comparable to each other. 

We also extract the MOS on evaluating the effectiveness of different models on controlling linguistic phonological symbols of synthesized speech using the JSUT dataset (Table~\ref{tab:tts_mos}). The evaluation was performed manually by 200 standard-Japanese speakers. Deep Voice 3~\cite{ping2018deep} beats two other models, Tacotron 2 and Transformer TTS, by 0.31 on average in this experiment. Deep Voice 3 is an attention-based model for text-to-speech transformation, which is also suitable to manage the sequential data of speech. The MOS score obtained from those three models also are close to each other, while two sequential models Deep Voice 3 and Transformer TTS perform slightly better.

\begin{table}
\parbox{.45\linewidth}{
\centering
\caption{MOS on audio quality with 95\% confidence intervals}
\begin{tabular}{c|c}
    \toprule
    \hline
         Model &  MOS (audio quality) \\\hline
         Tacotron 2 &3.86$\pm$ 0.09\\
         \hline
         Merlin &2.40$\pm$ 0.13\\
         \hline
         Transformer TTS & \textbf{3.88$\pm$ 0.09} \\
         \hline
         FastSpeech &3.84 $\pm$ 0.08 \\
         \hline
    \end{tabular}
    \label{tab:tts}
}
\hfill
\parbox{.45\linewidth}{
\centering
\caption{MOS on effectiveness of controlling linguistic phonological symbols}
\begin{tabular}{c|c}
    \toprule
    \hline
         Model &  MOS (audio quality) \\\hline
         Tacotron 2 &3.20\\
         \hline
         Deep Voice 3 &\textbf{3.56}\\
         \hline
         Transformer TTS & 3.31 \\
         \hline
    \end{tabular}
    \label{tab:tts_mos}
}
\end{table}

\section{Datasets}
\label{sec:data}

As controllable deep data generation covers a few data modalities ranging from graph, image, to text, audio, 3D point cloud, time series, and tabular data, in this section, we summarize datasets that have been employed for controllable deep data generation across various modalities in Table~\ref{tab:data} and introduce datasets from each Modality in detail as follows.
\hbadness=99999
\begin{table}[]
\caption{Representative datasets for controllable deep data generation in different modalities}
% \begin{adjustbox}{width=0.8\textwidth}
\begin{tabularx}{0.8\textwidth}{|l|l|X|}
\cline{2-3} \hline
Modality                 & Type              & Dataset                                              \\ \cline{2-3} \hline
\multirow{4}{*}{Gragh} & Molecule          & QM9~\cite{ramakrishnan2014quantum}, ZINC250K~\cite{irwin2012zinc}, ChEMBL~\cite{mendez2019chembl}, MOSES~\cite{moses2020}                                      \\ \cline{2-3} 
                       & Synthetic network & Waxman Graphs~\cite{guo2021deep}, Random Geometric Graphs~\cite{guo2021deep}, Erdos-Renyi Graphs~\cite{guo2019deep} \\ \cline{2-3} 
                       & Mesh              & MeshSeg~\cite{Chen2009meshseg}                                                    \\ \cline{2-3} 
                       & Crystal           & QMOF~\cite{rosen2021machine} \\ \cline{2-3} \hline
\multirow{6}{*}{Image} & Digit         & MNIST~\cite{lecun1998gradient}, The Street View House Numbers (SVHN)~\cite{netzer2011reading}\\ \cline{2-3} 
                       & Human character  & Sprites~\cite{li2018disentangled}, CelebFaces Attributes(CelebA)~\cite{liu2015faceattributes}, Flickr-Faces-HQ (FFHQ)~\cite{karras2019style}, CMU Multi-PIE face~\cite{gross2010multi}, Labeled Faces in the Wild (LFW)~\cite{LFWTech}, Helen~\cite{le2012interactive}, MetFaces~\cite{karras2020training}\\ \cline{2-3} 
                       & Animal  & Caltech-UCSD Birds 200~\cite{399}\\ \cline{2-3} 
                       & Shapes  & dSprites~\cite{dsprites17}, 3D shapes~\cite{3dshapes18}\\ \cline{2-3} 
                       & Fashion  & DeepFashion~\cite{liuLQWTcvpr16DeepFashion}\\ \cline{2-3} 
                       & General scene & ImageNet~\cite{deng2009imagenet}, LSUN~\cite{yu2015lsun}\\ \cline{2-3} \hline
\multirow{2}{*}{Text} & English  & Yelp~\cite{guu2018generating}, Amazon~\cite{mcauley2015image}, Wikipedia~\cite{keskar2019ctrl}, STS benchmark (STSb)~\cite{cer2017semeval}, One Billion Word~\cite{chelba2013one}, RottenTomatoes~\cite{ficler2017controlling}, VerbNet~\cite{schuler2005verbnet}, ConceptNet~\cite{speer2012representing}, Stanford Sentiment Treebank (SST)~\cite{socher2013recursive}, CAPTIONS~\cite{gan2017stylenet}, Project Gutenberg~\cite{gerlach2020standardized}, ViGGO corpus~\cite{juraska2019viggo}, ParaNMT-50M~\cite{wieting2017paranmt}, IMDB text corpus~\cite{diao2014jointly}, TimeBank~\cite{pustejovsky2006timebank}, Facebook politicians~\cite{li2018delete}, EMNLP2017 WMT News~\cite{guo2018long}, OpenWebText~\cite{radford2019language}, ROCStories~\cite{xu2020megatron}\\ \cline{2-3} 
                       & Chinese & Taobao~\cite{shao2021controllable}, Chinese Poems~\cite{zhang2014chinese}\\ \cline{2-3} \hline
\multirow{4}{*}{Audio} & English  & LJSpeech~\cite{ljspeech17}, LibriTTS~\cite{tits2019methodology}, LibriSpeech corpus~\cite{panayotov2015librispeech}, LibriVox~\cite{kearns2014librivox}, Emotional Speech Database~\cite{zhou2021seen}, Catherine Byers~\cite{prahallad2013blizzard}, Youtube~\cite{hsu2018hierarchical}, dMelodies~\cite{pati2020dmelodies}, The Interactive Emotional Dyadic Motion Capture (IEMOCAP)~\cite{busso2008iemocap}, The RECOLA Database~\cite{ringeval2013introducing}, CMU ARCTIC database~\cite{kominek2003cmu}, MAESTRO~\cite{hawthorne2018enabling}, VoxCeleb2~\cite{chung2018voxceleb2}, TED-LIUM 3~\cite{hernandez2018ted}\\ \cline{2-3}
                       & Chinese & Chinese professional actress~\cite{zhu2019controlling}, Baidu Speech Translation Corpus (BSTC)~\cite{zhang2021bstc}\\ \cline{2-3} 
                       & Japanese& JSUT Corpus~\cite{sonobe2017jsut}, Japanese Emotional Speech Database~\cite{mori2006emotional}\\ \cline{2-3} 
                       & French  &SIWIS French Speech Synthesis Database~\cite{goldman2016siwis}\\\cline{2-3} \hline
\multirow{1}{*}{Table} & \multicolumn{2}{l|}{LACity~\cite{park2018data}, Adult~\cite{park2018data}, Health~\cite{park2018data}, Airline~\cite{jacobini2020bureau}, GEFCom2012~\cite{moon2020conditional}}\\ \cline{2-3} \hline
\multirow{1}{*}{3D point} & \multicolumn{2}{l|}{ShapeNet~\cite{chang2015shapenet}, Cityscapes 3D~\cite{cordts2016cityscapes}}\\ \cline{2-3} \hline
\multirow{1}{*}{Time series} & \multicolumn{2}{l|}{M1~\cite{godahewa2021monash}, M3~\cite{godahewa2021monash}, M4~\cite{godahewa2021monash}, Tourism~\cite{athanasopoulos2011tourism}, NN5~\cite{taieb2012review}, Philips eICU~\cite{pollard2018eicu}}\\ \cline{2-3} \hline
\end{tabularx}
% \end{adjustbox}
\label{tab:data}
\end{table}

\subsection{Graph}

Datasets for controllable deep graph generation include those for molecule design, synthetic networks, mesh data generation and crystal structure design~\cite{du2021graphgt,huang2021therapeutics,jamasb2022graphein}. \textit{QM9} is an enumeration of around 134k stable organic molecules with up to 9 heavy atoms that have 17 chemical properties labelled in the dataset~\cite{ramakrishnan2014quantum}. QM9 has been broadly employed in tasks related to molecule synthesis and optimization~\cite{de2018molgan, ma2018constrained, madhawa2019graphnvp, liu2021graphebm, du2020interpretable, gebauer2019symmetry, zang2020moflow, gomez2018automatic}. Similarly, \textit{ZINC250K}~\cite{irwin2012zinc} is another dataset for molecule design that contains around 250k commercially available drug-like chemical compounds labeled by various physical and chemical properties~\cite{you2018graph, jin2018junction, ma2018constrained, li2018learning, putin2018reinforced, madhawa2019graphnvp, liu2021graphebm, du2020interpretable, shi2020graphaf, yang2020practical, zang2020moflow}. \textit{ChEMBL}~\cite{mendez2019chembl} is a manually curated database of bio-active molecules with drug-like properties and has been applied to multi-objective, optimization-based molecule design~\cite{xie2020mars, jin2020multi, wang2021multi}. In addition to molecular design, some works aim to generate graphs with specific geometric properties, which require synthesizing graphs with those properties as the training data. For instance, Waxman Graphs, Random Geometric Graphs and Erdos-Renyi Graphs have been generated by Guo et al.~\cite{guo2021deep, guo2019deep} to serve as the synthetic networks for this task. Moreover, \textit{MeshSeq}~\cite{Chen2009meshseg} is the dataset that contains 380 meshes for quantitative analysis of how people decompose objects into parts and for comparison of mesh segmentation algorithms~\cite{Chen2009meshseg}. This dataset has been borrowed to generate periodic graphs with different basic units~\cite{wang2022deep}. \textit{QMOF}~\cite{rosen2021machine} dataset is a publicly available database of computed quantum-chemical properties and molecular structures of metal–organic frameworks (MOFs)~\cite{rosen2021machine}. This dataset has also been used for controllable periodic graph generation~\cite{wang2022deep}.

\subsection{Image}

Datasets for image generation usually contain images and the corresponding labels. Handwritten digits are one significant type of images for visualizing controllable performance of deep learning model. For instance, the \textit{MNIST} dataset contains 70,000 images of handwritten digits with a $28 \times 28 \times 1$ array of floating-point numbers representing grayscale intensities ranging from 0 (black) to 1 (white) \cite{lecun1998gradient}. Each image of MNIST contains a one-hot binary vector of size ten indicating the digit categories of zero through nine. \textit{The Street View House Numbers (SVHN)} dataset includes over 600,000 labelled digit images \cite{netzer2011reading}, much more than those in MNIST, which is obtained from house numbers in Google Street View images. Another important and commonly used type of image data for controllable deep image generation is Sprites data. For example, \textit{Sprites} dataset contains 60 pixel color images of animated characters \cite{li2018disentangled}. Sprites contains 120,000 images in total, any of which has 7 sources of variation: body type, gender, hair type, armor type, arm type, greaves type, and weapon type \cite{mathieu2016disentangling}. \textit{dSprites} is another commonly used dataset of 2D shapes generated from 6 ground truth independent latent factors including color, shape, scale, rotation, x and y positions of a sprite \cite{dsprites17}. \textit{3D shapes} dataset contains 480,000 images regarding 3D shapes procedurally generated from 6 ground truth independent latent factors, including floor colour, wall colour, object colour, scale, shape and orientation \cite{3dshapes18}. Meanwhile, face data is one type of image data that is popular on image editing-related tasks. \textit{CelebFaces Attributes Dataset (CelebA)} is a large-scale face dataset with more than 200,000 celebrity images \cite{liu2015faceattributes}. CelebA contains 10,177 number of identities, 202,599 number of face images, and 5 landmark locations along with 40 binary attributes annotations per image. The \textit{Flickr-Faces-HQ} dataset consists of 70,000 high-quality images with considerable variation in terms of age, ethnicity, and image background. The FFHQ dataset was first employed to evaluate StyleGAN, which performs unsupervised separation of high-level attributes of face images \cite{karras2019style}. \textit{CMU Multi-PIE face} database contains more than 750,000 images of 337 people recorded in up to four sessions over the span of five months and labeled by the expression of the person in the image \cite{gross2010multi}. \textit{Labeled Faces in the Wild (LFW)} dataset contains 13,233 images of 5,749 people labelled with the individuals' names \cite{LFWTech}. \textit{Helen} dataset consists of 2,330 images constructed using annotated Flickr images \cite{le2012interactive}. Helen was hand-annotated using Amazon Mechanical Turk to locate precisely the eyes, nose, mouth, eyebrows, and jawline and originally intended for building a facial feature localization algorithm. \textit{MetFaces} dataset is an image dataset of human faces extracted from works of art, consisting of 1,336 high-quality images at $1024\times 1024$ resolution \cite{karras2020training}. \textit{Anime Face} dataset has 63,632 anime faces scraped from \url{www.getchu.com} but without labels in its original version. In addition to faces, \textit{Caltech-UCSD Birds 200} dataset contains 11,788 images annotated with 200 bird species \cite{399}. \textit{DeepFashion} database contains over 800,000 diverse fashion images annotated with 50 categories, 1,000 descriptive attributes, bounding boxes and clothing landmarks \cite{liuLQWTcvpr16DeepFashion}.

In addition to datasets that target specific types of objects, other datasets contain more general types of images. For instance, the \textit{MIRFLICKR-25000} dataset contains 25,000 annotated images downloaded from the social photography site Flickr that covers various topics \cite{huiskes08}. \textit{ImageNet Large Scale Visual Recognition Challenge (ILSVRC)} dataset evaluates algorithms for object detection and image classification on a large scale. ILSVRC consists of 150,000 photographs, hand labeled with the presence or absence of 1000 object categories that contain both internal nodes and leaf nodes of ImageNet. The \textit{CIFAR-10} dataset contains 60,000 $32\times32$ coloured images in 10 classes, with 6,000 images per class \cite{krizhevsky2009learning}. The label for each image in CIFAR-10 corresponds to image categories. \textit{LSUN} dataset contains around one million labeled images for each of 10 scene categories and 20 object categories \cite{yu2015lsun}. \textit{ImageNet} is an image dataset that populates the majority of the 80,000 synsets of WordNet with an average of 500-1000 clean, annotated and full resolution images \cite{deng2009imagenet}. The \textit{NORB} dataset is intended for experiments in 3D object recognition from shape \cite{lecun2004learning}. NORB stores 29,160 images of 50 toys belonging to 5 generic categories: four-legged animals, human figures, airplanes, trucks and cars. Similarly, \textit{smallnorb} dataset is from the same source of images and also intended for 3D object recognition from shape, but with only 24,300 image pairs \cite{lecun2004learning}. 

\subsection{Text}

Datasets for controllable deep text generation along with available annotations according to the need of various tasks usually are collected from the public websites. For instance, \textit{Yelp review} dataset that contains 6,990,280 records from reviewers with over 1.2 million business attributes serves as a popular dataset for controllable text-style transfer \cite{keskar2019ctrl, xu2020variational, yang2018unsupervised, john2018disentangled, sudhakar2019transforming} and prototype editing \cite{guu2018generating}. Another famous dataset in this domain is \textit{Amazon review} that has 142.8 million reviews with product metadata  such as descriptions, category information, price, brand, and image features \cite{mcauley2015image}. Amazon review dataset is also adopted in the task of text style transfer \cite{xu2020variational, li2018delete, sudhakar2019transforming, john2018disentangled} or specific tasks such as question answering and translation \cite{keskar2019ctrl}. \textit{Wikipedia} dataset contains 5,075,182 records in SQL file format and serves as a large corpus to train and evaluate the model for controllable deep text generation \cite{keskar2019ctrl, chang2021changing}. \textit{STS benchmark (STSb)} dataset was originally designed for the semantic textual similarity task \cite{cer2017semeval}. Since the sentences from STSb are easier to understand for annotators compared with Wikipedia, it has been adopted to predict upcoming topics \cite{chang2021changing}. The \textit{One Billion Word} dataset is another corpus that contains 0.8 billion words for language modeling \cite{chelba2013one}. This dataset was used for prototype editing tasks in the domain of controllable deep text generation \cite{guu2018generating}. The \textit{RottenTomatoes} dataset was collected by Ficler and Goldberg~\cite{ficler2017controlling}, in which 1,002,625 movie reviews for 7,500 movies were collected. Rottentomatoes was first employed on controlling Controlling the linguistic style of generated sentences \cite{ficler2017controlling}. Besides the well-established datasets, \textit{VerbNet} is the largest on-line English verb lexicon \cite{schuler2005verbnet}. Each verb class in VerbNet is described by thematic roles, selectional restrictions on the arguments, and frames consisting of a syntactic description and semantic predicates with a temporal function. VerbNet has been employed to train the model for controllable neural story plot generation \cite{tambwekar2018controllable}. Similarly, \textit{ConceptNet} is another semantic network that consists of 600k knowledge triples \cite{speer2012representing}. \textit{Stanford Sentiment Treebank} (SST) dataset includes fine-grained sentiment labels for 215,154 phrases in the parse trees of 11,855 sentences \cite{socher2013recursive}. SST was originally intended for testing NLP model’s abilities on sentiment analysis, but recently was also borrowed to control over a range of topics and sentiment styles of generated texts \cite{dathathri2019plug}. The \textit{CAPTIONS} dataset contains sentences describing 10,000 images labeled as either factual, romantic, or humorous \cite{gan2017stylenet}. CAPTIONS dataset has been borrowed in style transfer tasks \cite{li2018delete}. \textit{Project Gutenberg} contains more than 50,000 books and more than $3\times 109$ word-tokens \cite{gerlach2020standardized}. \textit{ViGGO corpus} is a set of 6,900 meaning representations to natural language utterance pairs in the video game domain \cite{juraska2019viggo}. ViGGO corpus has been borrowed to evaluate how the phrase-based data augmentation method can improve controllable deep text generation \cite{kedzie2020controllable}. \textit{ROCStories} includes 98,159 stories and have been used in controllable deep text generation by incorporating an external knowledge \cite{xu2020megatron}. \textit{ParaNMT-50M} is a dataset of more than 50 million English-English sentential paraphrase pairs that provide a rich source of semantic knowledge \cite{wieting2017paranmt}. \textit{IMDB text corpus} contains 350K movie reviews and has been used for controllable deep sentiment generation \cite{diao2014jointly, hu2017toward}. \textit{TimeBank} dataset contains 183 English news articles with over 27,000 events and temporal annotations, adding events, times and temporal links between events and times \cite{pustejovsky2006timebank}. Timebank was compiled in Hu et al.~\cite{hu2017toward} to extract a lexicon of 5250 words and labeled phrases for controllable deep text generation. \textit{Facebook politicians} dataset includes responses to Facebook posts from members of the U.S. House and Senate \cite{voigt2018rtgender}. This dataset contains 399,037 source texts, 13,866,507 responses and 376,114,950 word counts in total. Facebook politicians also was a part of the training model for style transfer purposes \cite{li2018delete}. The \textit{EMNLP2017 WMT News} dataset consists of 646,459 words and 397,726 sentences \cite{bojar2017findings}, and has been used for long text generation \cite{guo2018long}. \textit{OpenWebText} dataset contains over 10 million HTML pages where users can extract texts \cite{radford2019language} and has been used in controllable deep text generation regarding style, content, and task-specific behavior \cite{keskar2019ctrl}. 

In addition to English datasets, the language datasets based on other languages were also borrowed in controllable deep text generation tasks. For instance, Shao et al.~\cite{shao2021controllable} collected the Chinese dataset from \textit{Taobao}, a Chinese E-commerce platform to train the model controlling the order of the keywords of generated sentences. This dataset contains 617,181 items and 927,670 item text descriptions. The total size of the vocabulary used is 88,156. \textit{Chinese Poems} dataset contains 284,899 4-line 5-character Chinese poems in total \cite{zhang2014chinese} and has been borrowed in short text generation tasks \cite{shao2021controllable}. 

\subsection{Audio}

Datasets for controllable deep audio generation contain a range from English to other languages such as French, Japanese and Chinese. The \textit{LJSpeech} dataset consists of 13,100 short audio clips of a single speaker reading passages from 7 non-fiction books. Clips vary in length from 1 to 10 seconds and have a total length of approximately 24 hours \cite{ljspeech17}. LJSpeech dataset is commonly used in controllable TTS-related tasks \cite{tits2019exploring, ren2019fastspeech} or simple speech synthesis \cite{fabbro2020speech}. \textit{LibriTTS} dataset is a multi-speaker English corpus of approximately 585 hours of read English speech at 24kHz sampling rate and was specified employed in controllable TTS tasks \cite{valle2020mellotron, tits2019methodology}. \textit{LibriSpeech corpus} is a collection of approximately 1,000 hours of audiobooks that are a part of the LibriVox project. LibriSpeech also was borrowed in TTS-related task to control emotion of generated speech \cite{panayotov2015librispeech, tits2019methodology}.  \textit{LibriVox} is a collection of public audiobooks that can be used in controllable deep audio synthesis \cite{kearns2014librivox, sini2020introducing}. \textit{Emotional Speech Database} consists of 350 parallel utterances spoken by 10 native English and 10 native Chinese speakers and covers 5 emotion categories (neutral, happy, angry, sad, and surprised) \cite{zhou2021seen}. More than 29 hours of speech data were recorded in a controlled acoustic environment. ESD dataset has been serving for controllable deep emotional generation of speech \cite{liu2021reinforcement}. \textit{Catherine Byers} dataset was made available to registered participants in the Blizzard Challenge 2013 \cite{prahallad2013blizzard}. \textit{Youtube} serves as a resource for the community of the domain to sample audios per their needs \cite{hsu2018hierarchical}. \textit{dMelodies} dataset, which contains 2-bar monophonic melodies where each melody is the result of a unique combination of nine latent factors that span ordinal, categorical, and binary types, was originally intended for disentanglement learning on music generation \cite{pati2020dmelodies}. dMelodies was specifically used for controllable deep music generation \cite{pati2021disentanglement}. \textit{The Interactive Emotional Dyadic Motion Capture (IEMOCAP) database} is an acted, multimodal and multispeaker database that contains around 12 hours of audiovisual data, annotated by categorical labels, such as anger, happiness, sadness, neutrality, as well as dimensional labels such as valence, activation and dominance \cite{busso2008iemocap}. IEMOCAP has been borrowed for controllable deep emotion generation of speech \cite{cai2021emotion}. \textit{The RECOLA Database} consists of 9.5 hours of audio, visual, and physiological recordings of online dyadic interactions between 46 French speaking participants, who were solving a task in collaboration while affective and social behaviors were annotated \cite{ringeval2013introducing}. RECOLA was also used for controllable deep emotion generation on speech \cite{cai2021emotion}. The \textit{CMU ARCTIC database} consists of around 1150 utterances selected from out-of-copyright texts from Project Gutenberg \cite{kominek2003cmu}. The CMU ARCTIC database has been employed in the task of speech synthesis while controlling emotions \cite{tits2019methodology}. The \textit{MAESTRO} dataset contains around 200 hours of virtuosic piano performances captured with fine alignment between note labels and audio waveforms \cite{hawthorne2018enabling}. \textit{VoxCeleb2} contains over 1 million utterances for 6,112 celebrities, extracted from videos uploaded to YouTube \cite{chung2018voxceleb2}, and has been used to control the frequency of generated speech \cite{vasquez2019melnet}. \textit{TED-LIUM 3} is an audio dataset collected from TED Talks, containing 2,351 audio talks, 452 hours of audio and 2351 aligned automatic transcripts \cite{hernandez2018ted}. TED-LIUM 3 was similarly used to control the frequency of generated speech \cite{vasquez2019melnet}. 

In addition to English dataset, \textit{SIWIS French Speech Synthesis Database} includes 9,750 utterances from various sources such as parliament debates and novels uttered by a professional French voice talent \cite{goldman2016siwis}. The database includes more than ten hours of speech data and has been used in speech synthesis in controlling emotions \cite{goldman2016siwis}. \textit{Chinese professional actress} dataset mimics a little girl speaking in seven emotions (neutral, happy, angry, disgust, fear, surprise and sad) \cite{zhu2019controlling}. This dataset also has been used to generate audio with control on emotions \cite{li2021controllable}. \textit{JSUT Corpus} consists of 10 hours of reading-style speech data and its transcription and covers all the main pronunciations of daily-use Japanese characters \cite{sonobe2017jsut}. The JSUT dataset was used to prosodic features in speech generation \cite{kurihara2021prosodic}. \textit{Japanese Emotional Speech Database} contains 47 different ways of utterance to express 47 types of emotions including angry, joyful, disgusting, downgrading, funny, worried, gentle, relief, indignation, and shameful \cite{mori2006emotional}. \textit{Baidu Speech Translation Corpus (BSTC)} is a large-scale dataset for automatic simultaneous interpretation collected from the Chinese mandarin talks and reports \cite{zhang2021bstc}. This dataset contains 50 hours of real speeches, including three parts, the audio files, the transcripts, and the translations, and was borrowed for style control of generated speech \cite{bian2019multi}.

\subsection{Table}

Datasets that have been employed for controllable deep tabular data generation include the \textit{LACity} dataset that contains records of Los Angeles city government employees (such as salary, department and so on) \cite{park2018data}. \textit{Adult} dataset has many personal records (such as nationality, education level, occupation, work hours per week, and so forth) \cite{park2018data}. \textit{Health} dataset consists of various information (such as blood test results, questionnaire survey, diabetes, and so on) \cite{park2018data}. \textit{Airline} dataset selects $10\%$ out of all tickets sold in the USA and releases the data to the public every quarter \cite{jacobini2020bureau}. \textit{MNIST28} dataset was binarized by Xu et al.~\cite{xu2019modeling}, and each sample was converted to 784-dimensional feature vectors plus one label column to mimic high dimensional binary data. \textit{The Global Energy Forecasting Competition (GEFCom2012)} dataset includes hourly electric load data of a US utility and the temperature data from January 1, 2005 to December 31, 2008 for 20 zones, and was used for controllable deep table generation via conditional GANs \cite{moon2020conditional}. 

\subsection{3D point}

Datasets for 3D point generation include \textit{ShapeNet} that covers 55 common object categories with about 51,300 unique 3D models \cite{chang2015shapenet}. ShapeNet is widely used in 3D point generation tasks and serves as a popular dataset to control generated 3D point clouds. \textit{Cityscapes 3D} is an extension of the original Cityscapes with 3D bounding box annotations for all types of vehicles \cite{cordts2016cityscapes}. This dataset has been borrowed to address 3D reconstruction from a single image and its inverse problem of rendering an image given a point cloud \cite{pumarola2020c}. \textit{QM9} consists of around 134,000 organic molecules with up to nine heavy atoms from carbon, nitrogen, oxygen, to fluorine \cite{ruddigkeit2012enumeration}. QM9 was borrowed by Gebauer et al.~\cite{gebauer2019symmetry} to approximate molecules with 3D points and generate molecules with desired properties.

\subsection{Time series}

Datasets for controllable deep time series generation include \textit{M1} dataset that contains 1,001 time series data, \textit{M3} dataset that contains 3,003 time series data and \textit{M4} dataset that contains 100,000 data \cite{godahewa2021monash}, \textit{Tourism} that includes 366 monthly series, 427 quarterly series and 518 annual series \cite{athanasopoulos2011tourism}, \textit{NN5} that has 111 time series, representing about two years of daily cash money withdrawal amounts at ATM machines at one of the various cities in the UK \cite{taieb2012review}. \textit{Vehicle
and engine speed} dataset contains a set of signals, recorded in a fleet of 19 Volvo buses over a 3–5 year period \cite{parthasarathy2020controlled}. \textit{Philips eICU} dataset contains around 200,000 patients from 208 care units across the US, with a total of 224,026,866 entries divided into 33 tables \cite{pollard2018eicu}.

\section{Summary of applications}
The summary of applications and the related representative works can be found in Table~\ref{tab:application}.

\begin{table}[]
\caption{Applications of controllable deep data generation and representative works.}
% \begin{adjustbox}{width=0.8\textwidth}
\begin{tabularx}{\textwidth}{|l|l|X|}
\cline{2-3} \hline
Modality                 & Task              & Representative works                                              \\ \cline{2-3} \hline\hline
\multirow{2}{*}{Molecule} & Molecular synthesis & Attentive FP~\citep{xiong2019pushing}, HC-MLE~\citep{neil2018exploring}, RationaleRL~\citep{jin2020multi}, EDM~\citep{hoogeboom2022equivariant}, C5t5~\citep{rothchild2021c5t5} \\ \cline{2-3} 
                       & Molecular optimization & Mimosa~\citep{fu2021mimosa}, SCVAE~\citep{yu2022structure}, Mol-CycleGAN~\citep{maziarka2020mol}, Modof~\citep{chen2021deep}, GEGL~\citep{ahn2020guiding}, MolDQN~\citep{zhou2019optimization}, MSO~\citep{winter2019efficient}, JT-VAE~\citep{jin2018junction}, VJTNN~\citep{jin2018learning}  \\ \cline{2-3} \hline
\multirow{1}{*}{Protein} & Protein design  & AlphaFold2~\citep{jumper2021highly}, DL~\citep{wang2018computational}, Structured Transformer~\citep{ingraham2019generative}   \\ \cline{2-3} \hline
\multirow{2}{*}{Image} & Image synthesis  & DDPM~\citep{ho2020denoising}, ILVR~\citep{choi2021ilvr}, StyleGAN~\citep{karras2019style}, ControlGAN~\citep{li2019controllable}, Attribute-Decomposed GAN~\citep{men2020controllable}, 3D Generator~\citep{liao2020towards}, MS-COCO~\citep{hong2018inferring}, \\ \cline{2-3} 
                       & Image editing & Gan-control~\citep{shoshan2021gan}, CPCGAN~\citep{yang2021cpcgan}, CVAE~\citep{sohn2015learning}, CAGlow~\citep{liu2019conditional}, c-Glow~\citep{lu2020structured}, Faceshop~\citep{portenier2018faceshop}, Deep Plastic Surgery~\citep{yang2020deep}, Editgan~\citep{ling2021editgan}, Imagic~\citep{kawar2023imagic}   \\ \cline{2-3} \hline
\multirow{4}{*}{Text} & Text style transfer  & LM Classifier~\citep{yang2018unsupervised}, CTRL~\citep{keskar2019ctrl}, Cross-align~\citep{shen2017style}, DualRL~\citep{luo2019dual}, GST~\citep{sudhakar2019transforming}, Style transformer~\citep{dai2019style}   \\ \cline{2-3}
                       & Content creation & VRS~\citep{shen2019select}, Diffusion-lm~\citep{li2022diffusion}, MEGATRON-CNTRL~\citep{xu2020megatron}, DEXPERTS~\citep{liu2021dexperts}, SIC-Seq2Seq~\citep{luo2019learning}  \\ \cline{2-3} 
                       & Language translation& kNN-MT~\citep{khandelwal2020nearest}, dual-NMT~\citep{he2016dual},  GNMT~\citep{wu2016google}, RNN Encoder–Decoder~\citep{bahdanau2014neural}, Tensor2tensor~\citep{vaswani2018tensor2tensor}   \\ \cline{2-3} 
                       & Code generation & AlphaCode~\citep{li2022competition}, Codet~\citep{chen2022codet}, Codet5~\citep{wang2021codet5}, GPT-C~\citep{svyatkovskiy2020intellicode}   \\ \cline{2-3} \hline
\multirow{2}{*}{Audio} & Text-to-speech synthesis    & Fastspeech~\citep{ren2019fastspeech}, PromptTTS~\citep{guo2023prompttts}, Non-Attentive Tacotron~\citep{shen2020non}, Glow-TTS~\citep{kim2020glow}, Diff-TTS~\citep{jeong2021diff}   \\ \cline{2-3} 
                        & Music generation & Music
FaderNets~\citep{tan2020music}, Music SketchNet~\citep{chen2020music}, Theme transformer~\citep{shih2022theme}, Foley music~\citep{gan2020foley}   \\ \cline{2-3} \hline
\multirow{1}{*}{3D point} & 3D point cloud shape generation    & EditVAE~\citep{li2022editvae}, LAS-Diffusion~\citep{zheng2023locally}, CPCGAN~\citep{yang2021cpcgan}, SP-GAN~\citep{li2021sp}, SDF‐StyleGAN~\citep{zheng2022sdf}, LION~\citep{zeng2022lion}    \\ \cline{2-3} \hline
\multirow{1}{*}{Video} & Video generation     & TI2V~\citep{hu2022make}, CCVS~\citep{le2021ccvs}, ControlVideo~\citep{zhang2023controlvideo}, PC-AVS~\citep{zhou2021pose}     \\ \cline{2-3}
\multirow{1}{*}{Motion} & Motion generation     & Motiondiffuse~\citep{zhang2022motiondiffuse}, MultiAct~\citep{lee2023multiact}, MDM~\citep{tevet2022human}, TEMOS~\citep{petrovich2022temos}     \\ \cline{2-3}\hline
\end{tabularx}
% \end{adjustbox}
\label{tab:application}
\end{table}

\end{document}